\documentclass[sort&compress, 3p, number, a4paper]{elsarticle}

\usepackage[utf8]{inputenc}

\usepackage{amsmath}

\usepackage[T1]{fontenc}
\usepackage{lmodern}

\usepackage{hyperref} 
\usepackage{subfigure} 
\usepackage{multirow} 
\usepackage{array} 
\usepackage[modulo]{lineno} 
\usepackage{placeins} 

\begin{document}

\journal{Information Sciences}

\begin{frontmatter}

\title{Linguistic data mining with complex networks: a~stylometric-oriented~approach}

\author[label1]{Tomasz Stanisz}

\author[label1]{Jarosław Kwapień}

\author[label1,label2]{Stanisław Drożdż\corref{cor1}}

\address[label1]{Complex Systems Theory Department, Institute of Nuclear Physics, Polish Academy of Sciences, ul. Radzikowskiego 152, Kraków 31-342, Poland}

\address[label2]{Faculty of Physics, Mathematics and Computer Science, Cracow University of Technology, ul. Warszawska 24, Kraków 31-155, Poland}

\ead{stanislaw.drozdz@ifj.edu.pl}
\cortext[cor1]{Corresponding author}

\begin{abstract}
By representing a text by a set of words and their co-occurrences, one obtains a word-adjacency network being a reduced representation of a given language sample. In this paper, the possibility of using network representation to extract information about individual language styles of literary texts is studied. By determining selected quantitative characteristics of the networks and applying machine learning algorithms, it is possible to distinguish between texts of different authors. Within the studied set of texts, English and Polish, a properly rescaled weighted clustering coefficients and weighted degrees of only a few nodes in the word-adjacency networks are sufficient to obtain the authorship attribution accuracy over 90\%. A correspondence between the text authorship and the word-adjacency network structure can therefore be found. The network representation allows to distinguish individual language styles by comparing the way the authors use particular words and punctuation marks. The presented approach can be viewed as a generalization of the authorship attribution methods based on simple lexical features.

Additionally, other network parameters are studied, both local and global ones, for both the unweighted and weighted networks. Their potential to capture the writing style diversity is discussed; some differences between languages are observed.
\end{abstract}


\end{frontmatter}


\section{Introduction}

Many systems studied in contemporary science, from the standpoint of their structure, can be viewed as the ensembles of large number of elements interacting with each other. Such systems can often be conveniently represented by networks, consisting of nodes and links between these nodes. As nodes and links are abstract concepts, they may refer to many different objects and types of interactions, respectively. Therefore, a rapid development of the so-called \textit{network science} has found application in studies on a great variety of systems, like social networks, biological networks, networks representing financial dependencies, the structure of the Internet, or the organization of transportation systems \cite{Newman2003, Boccaletti2006, Drozdz2017, Valencia2009, Kwapien2012, Faloutsos1999, Sienkiewicz2005}.

The mentioned networks are usually referred to as complex networks; they owe their name to a general concept of \textit{complexity}. The above-listed systems are often viewed as examples of complex systems. Although complexity does not have a precise definition, it can be stated that a complex system typically consists of a large number of elements nonlinearly interacting with each other, is able to display collective behavior, and, by exchanging energy or information with surroundings, can modify its internal structure \cite{Kwapien2012}. Many complex systems exhibit emergent properties, which are characteristic for a system as a whole, and cannot be simply derived or predicted solely from the properties of the system's individual elements. The existence of the emergent phenomena is often summarized with the phrases such as "more is different" or "the whole is something besides the parts".

In this context, natural language is clearly a vivid instance of a complex system. Higher levels of its structure usually cannot be simply reduced to the sum of the elements involved. For example, phonemes or letters basically do not have any meaning, but the words consisting of them are references to specific objects and concepts. Likewise, knowing the meaning of separate words does not necessarily provide the understanding of a sentence composed of them, as a sentence can carry additional information, like an emotional load or a metaphorical message. Other features of natural language that are typical for complex systems have also been studied, for example long-range correlations \cite{Yang2016}, fractal and multifractal properties \cite{Ausloos2012, Drozdz2016, Chatzigeorgiou2017}, self-organization \cite{Steels2011, deBoer2011} or lack of characteristic scale, which manifests itself in power laws such as the well-known Zipf's law or Heaps' law (the latter also referred to as the Herdan's law) \cite{Yang2017, Piantadosi2014, Egghe2007}.

The network formalism has proven to be useful in studying and processing natural language. It allows to represent language on many levels of its structure - the linguistic networks can be constructed to reflect word co-occurrence, semantic similarity, grammatical relationships etc. It has a significant practical importance - the methods of analysis of graphs and networks have been employed in the natural language processing tasks, like keyword selection, document summarization, word-sense disambiguation or machine translation \cite{MIHALCEA_2011, Antiqueira2009, Amancio2011}. Networks also seem to be a promising tool in research on the language itself; they have been applied to study the language complexity \cite{Liu2008, Grabska-Gradzinska2012, Kulig2015}, the structural diversity of different languages \cite{Liu2011}, or the role of punctuation in written language \cite{Kulig2017}. Furthermore, the network formalism appears at the interface between linguistics and other scientific fields, for example in sociolinguistics, which, by studying social networks, investigates human language usage and evolution \cite{Kalampokis2007, DallAsta2006}.

In this paper the properties of the word-adjacency networks constructed from literary texts are studied. Such networks may be treated as a representation of a given text, taking into account mutual co-occurrences of words. They can be characterized by some parameters that describe their structure quantitatively. Determining these parameters results in obtaining a set of a few numbers capturing considered features of networks and, consequently, features of the underlying text sample. In that sense the procedure of constructing a network from a text and calculating its characteristics may be viewed as extracting essential traits of the language sample and putting them into a compact form. Subjecting the obtained data to data mining algorithms reveals that the networks representing texts of different authors can be distinguished one from another by their parameters. After appropriately proposed preparation of the data, such procedure of authorship attribution, though quite simple, results in relatively high accuracy. It therefore leads to a statement that information on individual language style used in a given text can be drawn from the set of characteristics describing the corresponding word-adjacency network. This is comparable with the known results concerning identification of individual language styles using the linguistic networks \cite{Mehri2012, Akimushkin2017}; however, the previous research has typically focused on the network types and the network structure aspects other than considered here.

\section{Data and methods \label{sec_data}}

\subsection{Word-adjacency networks}

The set of the studied texts consists of novels in English and Polish. For each language, there are 48 novels, 6 novels for each of 8 different authors. The full list of the books and authors is presented in \hyperref[appendix]{Appendix}. The texts were downloaded from the websites of Project Gutenberg \cite{PROJECT_GUTENBERG}, Wikisources \cite{WIKISOURCES}, and Wolne Lektury \cite{WOLNE_LEKTURY}. Both Polish and English-language authors considered in this paper are widely recognized as notable writers, whose works had an impact on the shaping of culture; two of the Polish writers - Henryk Sienkiewicz and Władysław Reymont - are laureates of the Nobel Prize in Literature. Among the selection criteria for the authors, worth mentioning are the number of books (each studied author has at least 6 books, each of them with at least 25000 words), and their availability (all the books are publicly available).

Each text studied in this paper is transformed into a word-adjacency network, and resulting networks are subjected to adequate calculations and analyses. However, before the construction of the networks, the texts are appropriately pre-processed. Preparation includes: removing annotations and redundant whitespaces, converting all characters to lowercase, and replacing some types of punctuation marks by special symbols, which are treated as usual words in the subsequent steps of analysis. This approach is justified by the observation of the fact that in terms of some statistical properties, punctuation follows the same patterns as words \cite{Kulig2017} and may be a valuable carrier of information on the language sample. The punctuation taken into consideration is: full stop, question mark, exclamation mark, ellipsis, comma, dash, colon, semicolon and brackets. The remaining punctuation types (for example, quotation marks) are removed from text samples, along with marks that do not indicate the structure of sentence (examples are: the full stops used after abbreviations, or quotation dashes - i.e.\ the dashes placed at the beginning of quotations in some languages, including Polish).

After pre-processing, the texts are transformed into the word-adjacency networks. In such networks each vertex represents one of the words appearing in the underlying text. The edges reflect the co-occurrence of words. In an unweighted network, an edge is present between two vertices, if the words that correspond to these vertices appear at least once next to each other in the text (see Figure \ref{fig_network_randomization}). In a weighted network, edges are assigned weights that represent the exact number of the co-occurrences of the respective words. It is assumed that the theoretically possible case of words occurring next to themselves is ignored - i.e.\ there are no edges connecting a vertex with itself in the resulting network.

An unweighted network may be treated as a special case of a weighted network (with all edge weights equal to 1). When determining the characteristics in a weighted network, one may either take weights into account or neglect them; the latter case is equivalent to the calculation of the parameters of an unweighted network. Therefore, all networks studied in this paper may be viewed as weighted ones. With such approach, weighted and unweighted versions of a given network characteristic can be treated as two distinct quantities describing a weighted network, with two different definitions. 

In the process of network construction, words are not lemmatized, which means that their inflected forms are distinguished. Hence all such forms occurring in a particular text end up as separate vertices in the resulting network. All studied networks are undirected - which means that a pair of adjacent words is counted in the same way as a pair with reversed order.

\subsection{Network characteristics}

Throughout this paper, a few concepts and quantities known from the field of complex network analysis are used. This section presents them briefly.

For each studied network, a number of characteristics are calculated. Depending on the context, these characteristics are either global (related to the network as a whole), or local (related to particular vertices in the network). The quantities describing the network's structure can be unweighted or weighted - the latter take into account the weights assigned to edges in the weighted networks. The characteristics studied in this paper are: vertex degree, average shortest path length, clustering coefficient, assortativity coefficient, and modularity. Although these quantities are widely applied in the studies on complex networks and are well described in literature \cite{F.Costa2007, Saramaeki2007}, there are some ambiguities, especially pertaining to generalizations onto the weighted networks. For this reason, the adequate definitions are given below.

\subsubsection{Vertex degree \label{subsec_degree}}

Degree is a quantity describing individual vertices. In simple (i.e.\ unweighted) networks, the degree of a given vertex $v$ is the number of edges touching $v$, and is denoted by $\deg(v)$. In weighted networks, the degree can be generalized to weighted degree (also called \textit{strength} and therefore denoted by $\text{str}(v)$), which is the sum of weights of the edges touching $v$.

In word-adjacency networks, the vertex degree is obviously related to the frequency of the corresponding word in the considered text sample. In its weighted version, it is equal to $2f(v)$, where $f(v)$ stands for the overall number of occurrences of the word $v$ (apart from the first and the last word in a sample, for which the weighted degree is equal to $2f(v)-1$). In unweighted networks, the degree of a vertex represents the number of words that appear at least once next to the word assigned to this vertex.

\subsubsection{Clustering coefficient \label{subsec_clustering_coeff}}

In unweighted networks, the clustering coefficient of a given vertex represents the probability that two randomly chosen direct neighbours of this vertex are also direct neighbours of each other. By a \emph{direct neighbour} of a vertex $v$ we understand a vertex connected with $v$ by an edge. Let $m_{v}$ stand for the number of edges in the network that link the direct neighbours of $v$ with other direct neighbours of $v$. Then the clustering coefficient of $v$ is given by:
\begin{equation}
C_u(v) = \frac{2m_{v}}{\deg(v)\cdot(\deg(v)-1)},
\end{equation}
where the subscript "$u$" comes from the word "unweighted". In a word-adjacency network, the clustering coefficient of a given word $v$ describes the probability that any two words that appear next to $v$ in the considered text sample also appear next to each other at least once.

Generalization of the clustering coefficient onto the weighted networks can be done in multiple ways. In this paper a definition proposed by Barrat et al. \cite{Barrat2004} is used. Let $S(v)$ denote the set of neighbours of a vertex $v$, and let $w_{uv}$ denote the weight of the edge connecting vertices $u$ and $v$ (if there is no such edge, then $w_{uv}=0$). Let $a_{uv}$ denote an unweighted adjacency matrix element, i.e.\ a number defined as follows:
\[
  a_{uv} =
  \begin{cases}
                                   1, & \text{if there exists an edge connecting $u$ and $v$;}\\
                                   0, & \text{otherwise.}
  \end{cases}
\]
Then the weighted clustering coefficient of $v$ is written as:
\begin{equation}
C_w(v) = \frac{1}{\text{str}(v) \cdot \left(\deg(v)-1\right)} \sum_{u,t \in S(v)} \frac{w_{vu} + w_{vt} }{2} a_{vu}a_{ut}a_{tv},
\end{equation}
where summation is over all pairs $(u,t)$ of neighbours of $v$. It is worth noting that if $\deg{v}=0$ or $\deg{v}=1$, the clustering coefficient cannot be determined from the above-given formulas, and it is assumed that in such cases it is equal to 0.

The above definitions pertain to the individual vertices of a network. Global clustering coefficient can be defined in more than one way; in this work a simple approach based on averaging the local clustering coefficients is applied. If $V$ stands for the set of all vertices of a network and $N$ is the number of elements in $V$, then the global clustering coefficient of the network is given by:
\begin{equation}
C = \frac{1}{N} \sum_{v \in V} C(v).
\end{equation}
Here the subscript "$u$" or "$w$", indicating the unweighted or weighted network, is omitted, because the formula is identical in both cases.

\subsubsection{Average shortest path length}

In unweighted networks, the length of a path between two vertices is the number of edges on that path. In weighted networks, by the length of a path we understand the sum of the reciprocals of edge weights on that path. The length of the shortest path between vertices $u$ and $v$ is also called the \emph{distance} between $u$ and $v$ and is denoted by $d(u,v)$.

The average shortest path length $\ell(v)$ of a vertex $v$ is the average distance from $v$ to every other vertex in the network. It is a measure of the centrality of a vertex in the network, and is given by the formula:
\begin{equation}
\ell(v) = \frac{1}{N-1} \sum_{ u \in V\backslash \{v\}}{d(v,u)} ,
\label{eq_mean_shortest_path_1}
\end{equation}
in which $V$ is the set of all vertices of the network, and $N$ is the number of elements in $V$.

The quantity defined above has finite values only in connected networks. If there are at least two vertices that are not connected by any path, the distance between them is not defined; usually it is treated as infinite, and therefore $\ell(v)$ cannot be calculated.

Global average shortest path length is a quantity describing the whole network; it is the average distance between all pairs of vertices. If local average distances $\ell(v)$ for all $v$ in $V$ are given, then the global mean distance in the whole network can be expressed by:
\begin{equation}
\ell = \frac{1}{N} \sum_{v \in V} \ell(v).
\label{eq_mean_shortest_path_2}
\end{equation}

Formulas \ref{eq_mean_shortest_path_1} and \ref{eq_mean_shortest_path_2} apply both to unweighted and weighted networks; the difference between the unweighted and the weighted average shortest path length lies in the definition of the distance in respective cases.

\FloatBarrier

\subsubsection{Assortativity coefficient}

Assortativity is a global characteristic of a network, describing the preference for vertices to attach to others that have similar degree. A network is called assortative, if vertices with high degree tend to be directly connected with other vertices with high degree, and low-degree vertices are typically directly connected to vertices which also have low degree. In disassortative networks, the high-degree nodes are typically directly connected to the nodes with low degree. In word-adjacency networks, assortativity provides the information about the extent to which frequent words co-occur with rare ones.

In unweighted networks, the assortativity coefficient can be defined as the Pearson correlation coefficient between the degrees of nodes that are connected by an edge. Let $(u, v)$ denote the ordered pair of vertices that are connected by an edge. Since edges are undirected and the pair $(u, v)$ is ordered, two such pairs can be assigned to each edge in the network. For each pair one can calculate the degrees of vertices $u$ and $v$, and form a pair $\left(\deg(u), \deg(v)\right)$. The set of all pairs $\left(\deg(u), \deg(v)\right)$ for all edges can be treated as the set of values of a certain two-dimensional random variable $(X,Y)$. With such notation, the assortativity coefficient $r_u$ is expressed by the Pearson correlation coefficient of variables $X$~and~$Y$:
\begin{equation}
r_u = \text{corr}(X,Y).
\end{equation}

The generalization of the above formula to weighted networks is done in this paper by replacing the degrees of vertices by their strengths, and calculating the weighted correlation coefficient instead of normal one. Let $(X,Y)$ be a two-dimensional random variable whose values are pairs $(x,y)=\left(\text{str}(u), \text{str}(v)\right)$ for all pairs of vertices $(u, v)$ connected by an edge. Let $w$ be a function that to each pair $(x,y)=\left(\text{str}(u), \text{str}(v)\right)$ assigns the weight of an edge connecting $u$ and $v$. Then the weighted assortativity coefficient $r_w$ can be written as:
\begin{equation}
r_w = \text{wcorr}(X,Y;w),
\end{equation}
where $\text{wcorr}(X,Y;w)$ denotes the Pearson weighted correlation coefficient of variables $X$ and $Y$ with the weighing function $w$. This formula is equivalent to the one that can be found, for example, in \cite{Leung2007}.

Since the assortativity coefficient is expressed by the correlation coefficient, it has values between -1~and~1. Networks with positive $r$ are assortative, while networks with negative $r$ are disassortative.

\begin{figure}[!ht]
\centering
\subfigure[]{\includegraphics[width=0.25\textwidth]{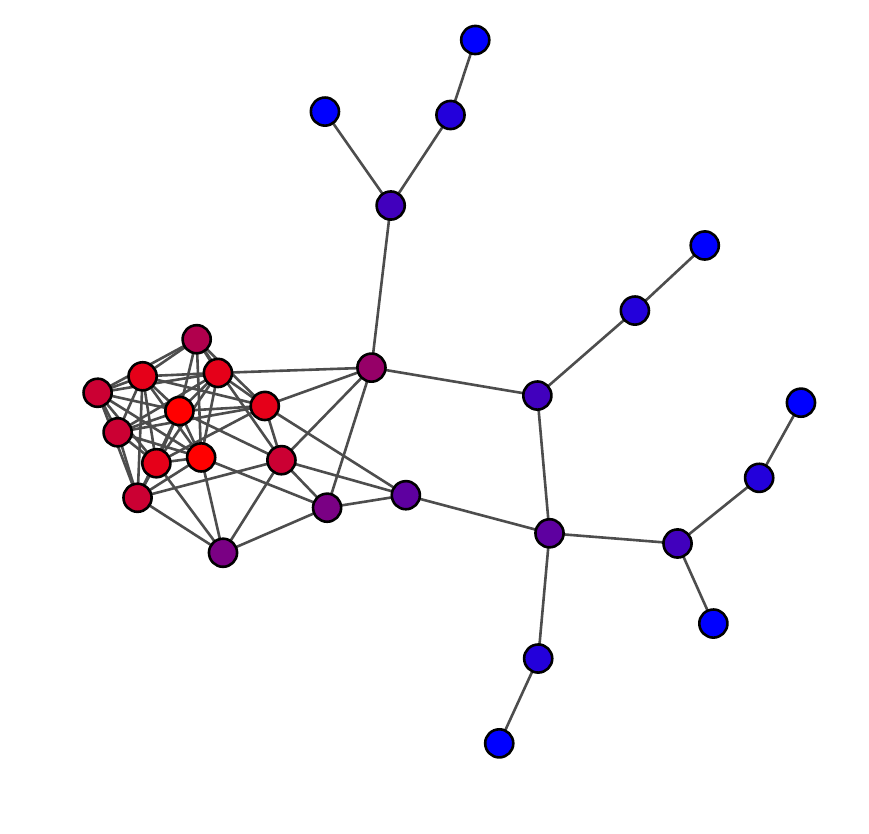}}
\qquad
\subfigure[]{\includegraphics[width=0.25\textwidth]{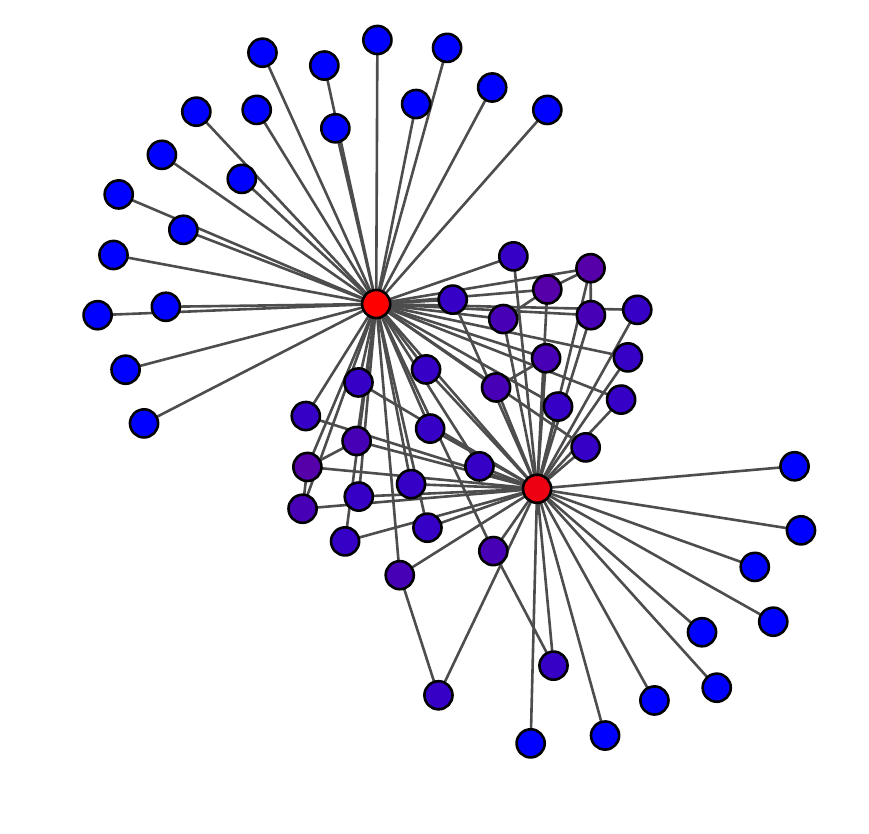}}
\qquad
\subfigure[]{\includegraphics[width=0.25\textwidth]{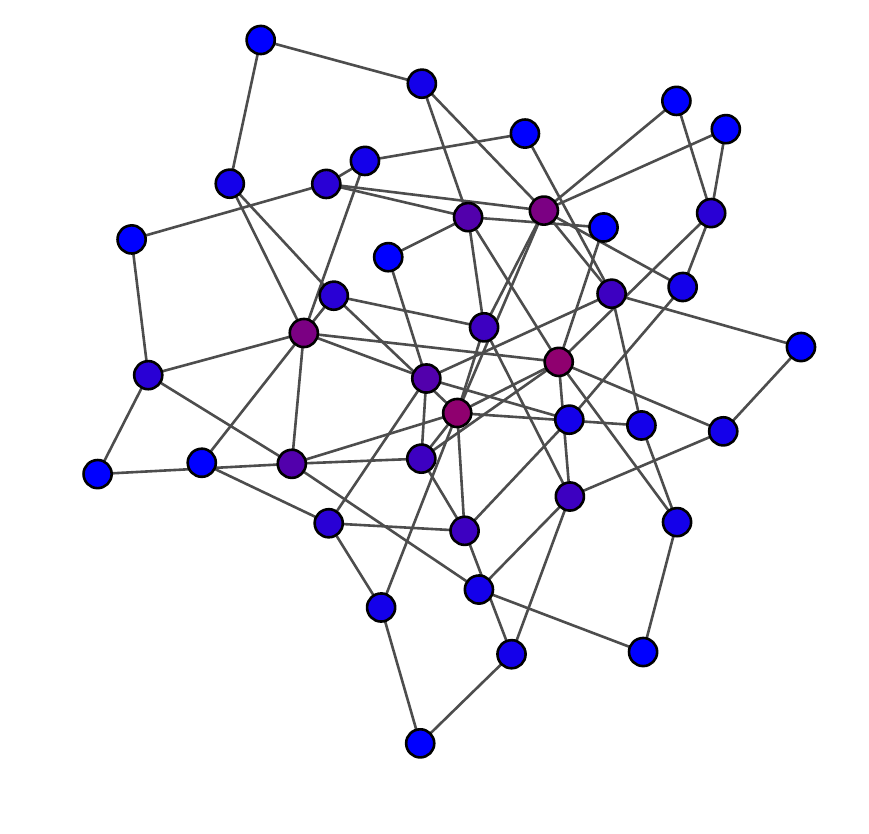}}
\caption{Unweighted networks with various assortativity coefficients. The network in (a) is assortative ($r=0.74$), the network in (b) is dissasortative ($r=-0.82$), and in the network in (c), the degrees of directly linked vertices are not correlated ($|r|<0.01$). In each network, the vertices are colored according to their degree - blue, purple and red colors correspond respectively to low, medium and high degree.}
\label{fig_assortativity}
\end{figure}

\subsubsection{Modularity \label{subsec_modularity}}

Modularity is a global characteristic of a network, measuring the extent to which the set of network's vertices can be divided into disjunctive subsets which maximize the density of edges within them, and minimize the number of edges connecting one with another. In word-adjacency networks, it can be interpreted as divisibility of the set of words appearing in the text into groups of words that frequently co-occur with each other. 

Consider an unweighted network, with the set of vertices $V$. By \textit{partition of the network} we understand the division of $V$ into disjoint subsets (called modules, clusters or communities). Let $a_{uv}$ denote the adjacency matrix element (defined the same as in subsection \ref{subsec_clustering_coeff}). Let $c_{v}$ denote the module to which the vertex $v$ is assigned by the given partition. The \textit{modularity of a partition} is defined as:
\begin{equation}
q_u = \frac{1}{2m} \sum_{u,v \in V} \left( \left[ a_{uv}-\frac{\deg(u) \deg(v)}{2m} \right] \delta(c_u,c_v) \right), \label{eq_modularity}
\end{equation}
where $m$ is the number of the edges of the network, $\deg(u),\deg(v)$ are degrees of vertices $u$ and $v$, and function $\delta(c_u,c_v)$ has value 1 if $c_u=c_v$ and 0 otherwise.

Modularity of a partition has value between -1 and 1, and indicates whether the density of edges within the given groups is higher or lower than it would be if edges were distributed at random. The random network that serves as a reference in this definition can be constructed using the so-called \emph{configuration model} \cite{Hofstad2016}.

The \textit{modularity of a network} $Q_u$ is the maximum value from modularities $q_u$ of all possible partitions. Determining the network's modularity precisely is computationally intractable, hence a number of heuristic algorithms have been proposed. In this paper modularity is calculated using the Louvain method \cite{Blondel2008}.

The generalization of modularity onto weighted networks can be done by replacing the quantities appearing in Eq. \ref{eq_modularity} by their weighted counterparts. If $w_{uv}$ denotes the weight of the edge connecting vertices $u$ and $v$, $M$ is the sum of all edge weights, and $\text{str}(u)$, $\text{str}(v)$ are the strengths of vertices $u, v$, then the modularity of a given partition is equal to:
\begin{equation}
q_w = \frac{1}{2M} \sum_{u,v \in V} \left( \left[ w_{uv}-\frac{\text{str}(u) \, \text{str}(v)}{2M} \right] \delta(c_u,c_v) \right).
\end{equation}
Again, the weighted modularity of the network, $Q_w$, is the greatest of modularities obtained in all possible partitions of the network.

\begin{figure}[!ht]
\centering
\subfigure[]{\includegraphics[width=0.28\textwidth]{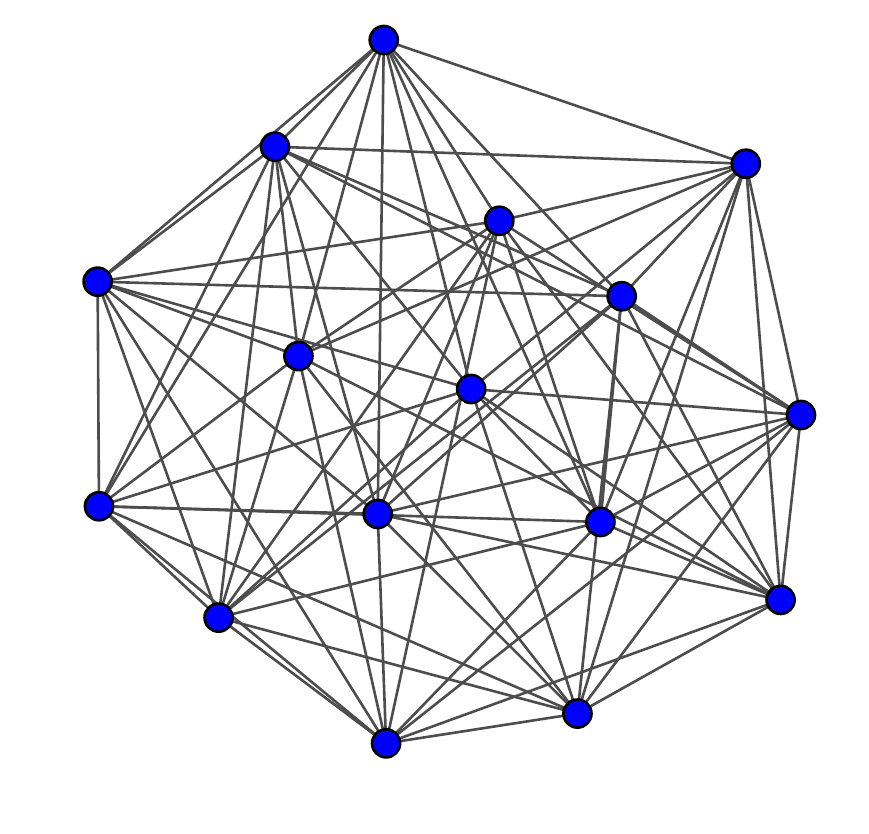}}
\qquad \qquad
\subfigure[]{\includegraphics[width=0.28\textwidth]{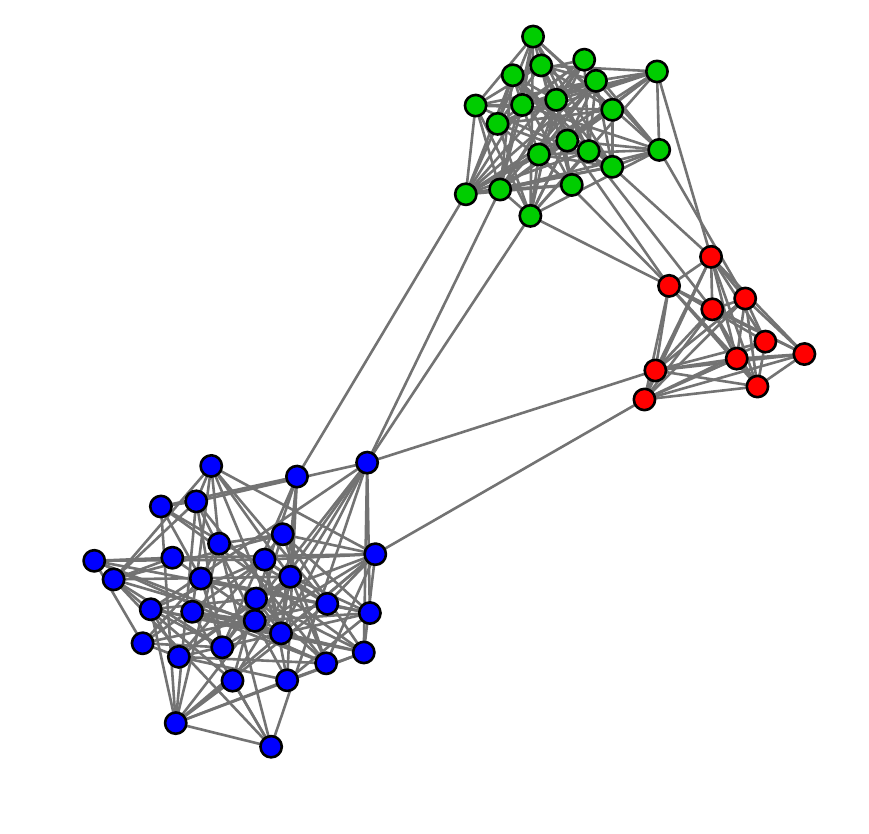}}
\caption{Examples of unweighted networks with (a) low modularity ($Q_u=0.07$) and (b) high modularity ($Q_u=0.58$). In figure b) the colors represent the partition which leads to the given value of modularity.}
\label{fig_modularity}
\end{figure}

\subsection{Characteristics' normalization \label{sec_normalization}}

\begin{figure}[t]
\centering
\subfigure[]{\includegraphics[width=0.45\textwidth]{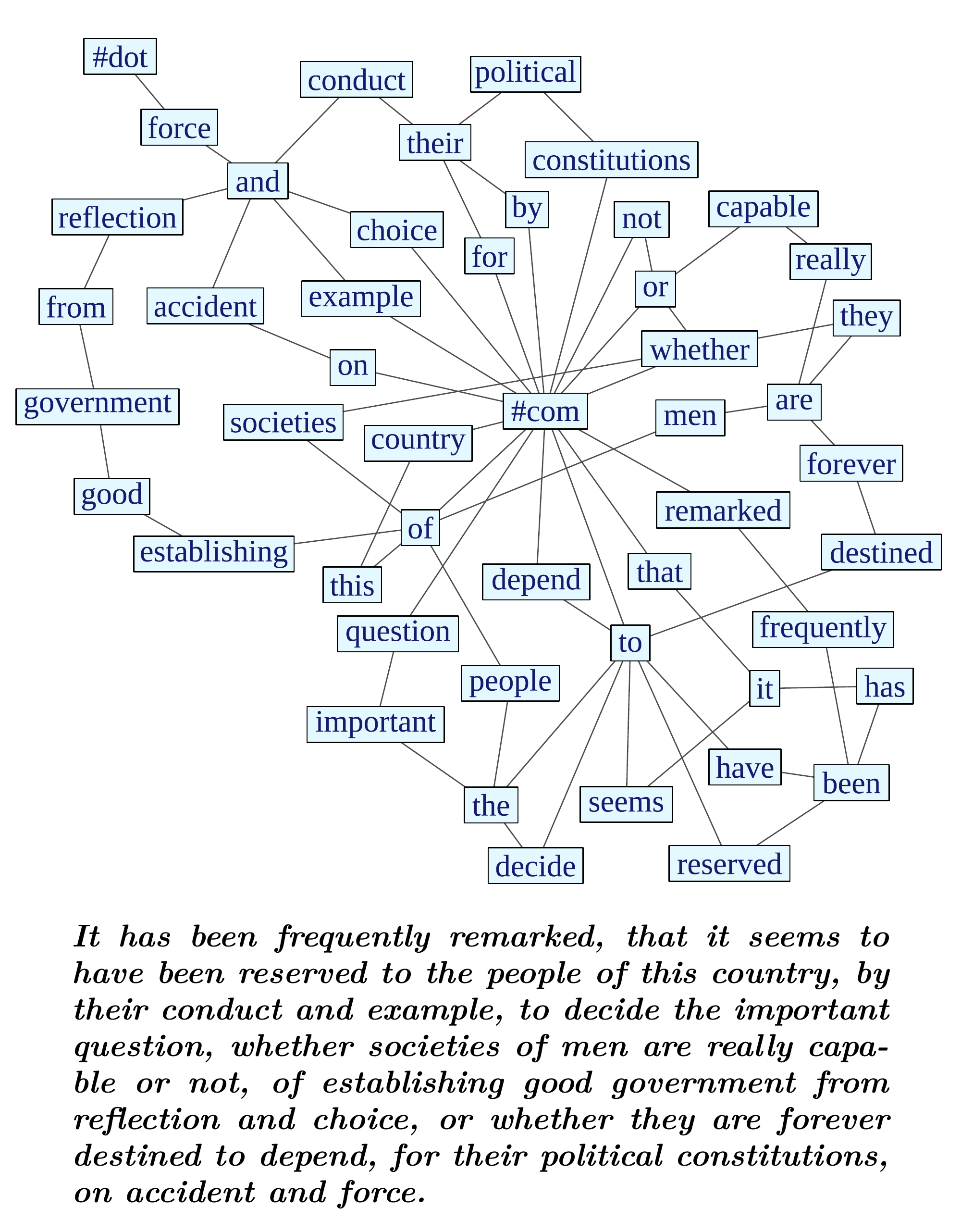}}
\quad
\subfigure[]{\includegraphics[width=0.45\textwidth]{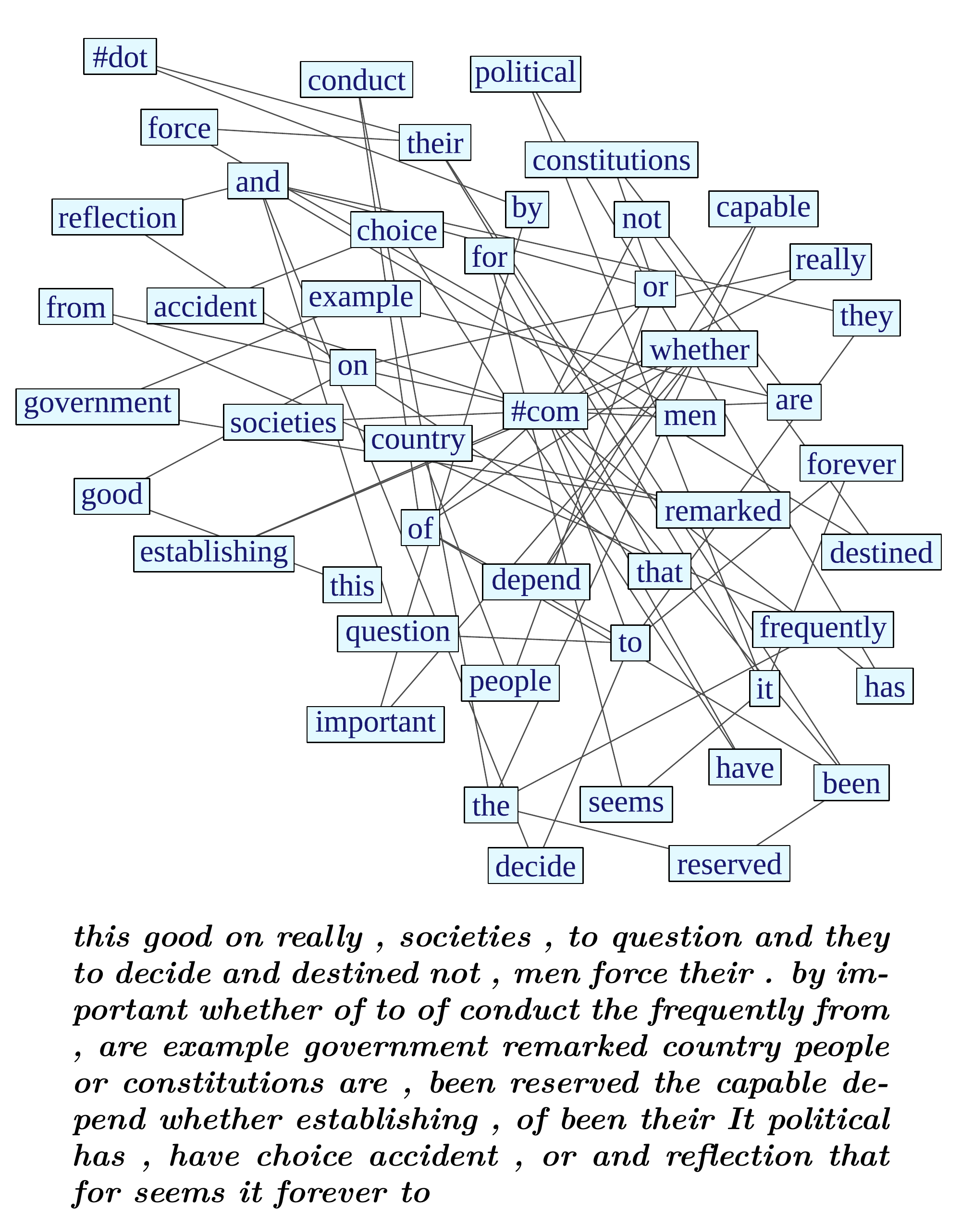}}
\caption{An unweighted word-adjacency network and its randomization. Figure (a) presents the network created from one sentence excerpted from \emph{The Federalist Paper No.\ 1} by Alexander Hamilton. Figure (b) shows the network created from the same piece of text, but with randomly shuffled words. Performing such shuffling repeatedly leads to obtaining a set of randomized networks which serve as a benchmark in the calculation of normalized characteristics of the original network. Note that the punctuation is taken into consideration in the construction of network; comma and full stop are denoted by "\#com" and "\#dot", respectively.}
\label{fig_network_randomization}
\end{figure}

The studied texts vary in length. To allow for a reasonable comparison between them, all calculated quantities are specifically normalized. Normalization of each network characteristic is achieved by dividing the value of the considered parameter by the average value of the same parameter in a network constructed from the randomly shuffled text (see Figure \ref{fig_network_randomization}). As an example, consider any text and any of above-mentioned quantities \ref{subsec_degree}-\ref{subsec_modularity}, say, the assortativity coefficient (it does not matter whether the weighted or the unweighted version is considered). To obtain the normalized (in described sense) value of the assortativity coefficient related to the given text, one needs to create the word-adjacency network from the text, and calculate the assortativity coefficient of this network, $r$. Then the original text needs to be subjected to the random shuffling of words, $k$ times (in this study, $k=50$ was used). From each of the $k$ obtained "random texts", a network can be created, in the same manner as in the previous step. Determining the assortativity coefficients for all of these networks and calculating their average gives $r'$, the typical value of the assortativity coefficient in the randomized text. The normalized value of the considered parameter - in this case, the assortativity coefficient - is then $\widetilde{r} = r/r'$. It reflects how the parameter related to the original text is different from the one obtained for a text with the same word distribution, but random word order. The normalization is performed in the described way for all mentioned characteristics \ref{subsec_degree}-\ref{subsec_modularity}. Normalized quantities are denoted with the symbol ''\textasciitilde '' throughout this paper.

\subsection{Data grouping and classification methods}

To explore the diversity of investigated networks' properties, two methods of data analysis were employed in this paper - hierarchical clustering and decision tree bagging. The former is a simple data clustering method, which, being an example of an unsupervised machine learning technique, attempts to identify the internal structure of a given set of data, based only on the data itself and on the assumed measure of similarity. The latter is a supervised learning method based on ensembles of decision trees.

\subsubsection{Hierachical clustering}

Hierarchical clustering \cite{Tan2005} performed on a given set of vectors, aims to group together the vectors that are close to each other, according to a certain metric (a function that defines the distance between vectors). The method, as used in this paper, works as follows: given an $m$-element set of  $n$-dimensional vectors of real numbers and a metric (here the Euclidean metric was used), it starts from creating $m$ clusters and assigning one vector to each cluster. Then the clusters that are closest to each other are merged into one, and this is repeated until there is only one cluster. As clusters are not vectors, but sets of vectors, the distance between such sets must be defined, to allow for determining which are closest to each other. This can be done in multiple ways, here the so-called furthest neighbour method was applied. It defines the distance between two clusters as the greatest distance between pairs of elements of these clusters (each element in a pair belongs to other cluster).

The result of clustering can be presented in the form of a dendrogram - a tree-like diagram which shows the consecutive merges and allows to determine what kind of clusters can be discerned within the dataset.

\subsubsection{Decision tree bagging}

Statistical classification with decision trees \cite{Sutton2005} is an example of supervised learning, which means that it allows for the categorization of data, provided that it is given a set of already categorized examples, called the training set.

Creating an ensemble of decision trees requires constructing individual trees in the first place. Let $A$ denote the set of $n$-dimensional vectors of real numbers; elements of $A$ are called observations, and their coordinates are called attributes. Each observation is labeled with one of $K$ categories, also called classes. The training of a single classification tree consists of: considering all possible \textit{one-dimensional splits} of $A$, selecting and executing the \textit{best split}, and repeating these steps recursively in the resulting subsets; splitting stops when $A$ is partitioned in such a way that each subset contains observations of only one category. The \textit{one-dimensional split} related to some attribute $x_i$ is the choice of a constant number $S$ and grouping the observations according to whether their coordinate $x_i$ is smaller or greater than $S$. The \textit{best split} is the one that maximizes the decrease of the diversity of distribution of classes in the considered set. The diversity can be measured, for example, by information entropy: $H=-\sum_{k=1} ^{K} p_k \log p_k$ ($p_k$ denotes the fraction of the observations in the set that belong to category~$k$). In such case, the maximization of diversity decrease is equivalent to the maximization of the quantity $H_0-H_{split}$, where $H_0$ is the initial entropy, and $H_{split}$ is the weighted sum of entropies in the resulting subsets, with weights proportional to the numbers of elements in these subsets and adding up to 1.

The scheme of the consecutive splits of $A$ is equivalent to a system of conditions imposed on the observations' attributes; such system is a classification tree. A trained tree can be used to categorize observations with unknown class membership, by assigning them to appropriate subsets of $A$, according to the conditions fulfilled by their components.

Classification with a single decision tree may suffer form instability, which means that the classifier may produce significantly different results for only slightly different training sets. Also, decision trees are prone to overfitting, which leads to decrease of classification accuracy of unknown observations.

Decision tree bagging (\emph{bootstrap aggregating}) \cite{Sutton2005} is a method of enhancing the performance of classification based on decision tress. Given a training set with $m$ observations, one can create $N$ new training sets of size $m$, by sampling with replacement from the original set. Obtained sets, called bootstrap samples, for large $m$ are expected to have the fraction $1-1/e$ (which is roughly 63.2\%) of the unique observations from the original set, the rest being duplicates. A decision tree is trained on each of the bootstrap samples, and the ensemble of $N$ trees becomes a new classifier. When such an ensemble is given an observation to classify, each tree being its part classifies the observation on its own, and then the class that was chosen by most of trees becomes the final result of classification.

A typical method of verifying the performance of classification is cross-validation \cite{Arlot2010}. Its general idea is dividing the set $A$ of observations with known class membership into two disjoint sets: the training set $A_{train}$ and the test set $A_{test}$. The classifier is trained on $A_{train}$ and then it classifies the observations in $A_{test}$, treating them as if their class memberships were unknown. Then the results are compared with the true class memberships of elements of $A_{test}$, and the number of correct matches indicates the classifier's performance. Partitioning $A$, training the classifier, and testing its performance is repeated a certain number of times, and the average result becomes the final assessment of classification's accuracy. The methods and rules of partitioning $A$ may vary; the approach utilized in this paper is the repeated random sub-sampling validation (also called Monte Carlo validation), with stratification. It simply performs an independent random stratified partition in each iteration, and each partition obeys a fixed proportion of the numbers of elements in the training and test set. Stratification ensures that all classes are equally represented in the training set.

\subsection{Computational issues}
The computational effort needed to perform the classification depends on the choice of classifiers' type, which is not necessarily restricted to decision trees and their ensembles. Analysis of networks, on the other hand, is based on algorithms whose exact running time can be found in the literature; it may depend on the type of network (whether it is a sparse network, for example) and its representation (adjacency matrix, adjacency list, etc.). However, it is worth mentioning which stages are crucial for the overall processing time. For networks of sizes as studied in this paper, the only noticeably time-consuming part of network characteristics' computation is determining the global average shortest path length, as it requires finding all shortest paths in a graph. To illustrate with an example: for the network constructed from Charles Dickens' \textit{David Copperfield} (the longest English-language book considered in the analysis) computing all network parameters that do not rely on determining shortest paths (namely, all vertex degrees and clustering coefficients, assortativity, modularity) both in unweighted and weighted version, takes 1.8 s on the PC used to perform the presented analysis (2.7 GHz CPU, 12 GB RAM). Obtaining average shortest path length (local) of the nodes corresponding to 100 most frequent words requires 0.1 s and 0.8 s, for the unweighted and weighted network, respectively. But getting global average shortest path length takes 12.3 s for unweighted, and 141.3 s for the weighted network. Considering that the characteristics' normalization requires repeating the computations for a number of randomized networks, it makes analysing the global shortest-path-related properties of networks rather impractical. However, as shown below, it turns out that global characteristics are not as effective at distinguishing between authors as local ones; furthermore, among studied network characteristics, these related to the lengths of shortest paths seem to have the weakest discriminative potential. Therefore, for practical purposes, they can be omitted without the loss of accuracy of the proposed approach.

\FloatBarrier
\section{Results}

\subsection{Global characteristics of unweighted networks}

\begin{figure}[htbp]
\centering
\includegraphics[width=0.95\textwidth]{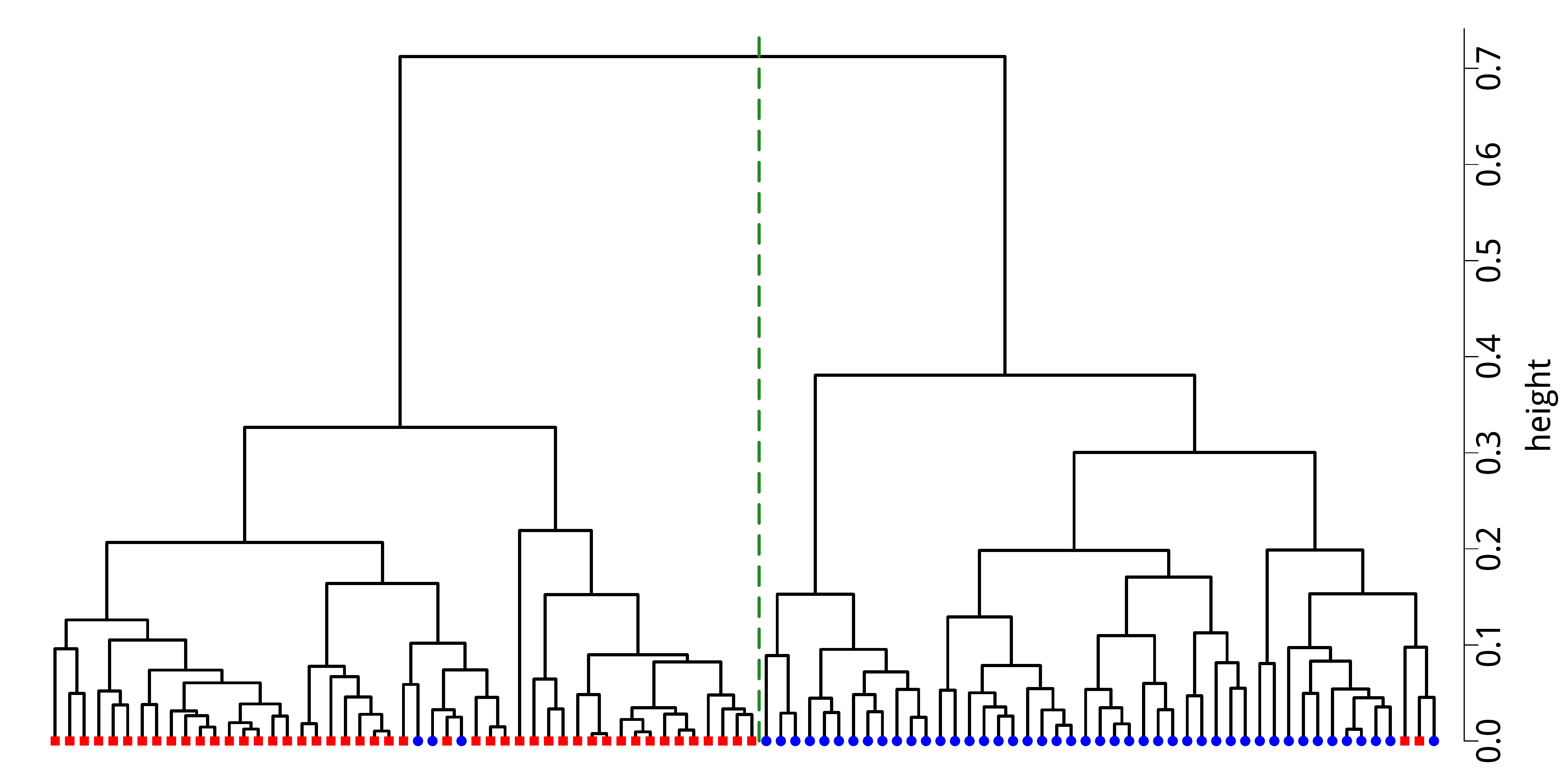}
\caption{The dendrogram of hierarchical clustering of all books listed in \hyperref[appendix]{Appendix}, in the space of \textbf{unweighted global characteristics of networks}. Red squares and blue dots denote English and Polish texts, respectively. The dashed line separates two general clusters, corresponding to distinct languages. The books falling into the "wrong" clusters are: \emph{Faraon}, \emph{Lalka}, \emph{Emancypantki} (Polish), \emph{Animal Farm}, \emph{The Prince and the Pauper} (English).}
\label{fig_dendro_all}
\end{figure}

At the beginning, the unweighted networks are studied, as they are simpler and less computationally demanding than the weighted ones. Each of the books listed in \hyperref[appendix]{Appendix} is converted into a network. Then the unweighted global parameters are calculated (all normalized, as described in \ref{sec_normalization}): average shortest path length $\widetilde{\ell}_u$, clustering coefficient $\widetilde{C}_u$, assortativity coefficient $\widetilde{r}_u$ and modularity $\widetilde{Q}_u$. The obtained sets of quantities are treated as vectors in 4-dimensional Euclidean space; each vector corresponds to one text, and each parameter corresponds to one component of a vector. The Euclidean distance is introduced as the measure of similarity (the greater the distance, the lesser the similarity), and hierarchical clustering is applied. The result indicates clear separation of the books with respect to the language (Figure \ref{fig_dendro_all}). The books cluster into two groups, each related to one language; only a few of them fall into the "wrong" group.

\begin{figure}[htb]
\centering
\subfigure[]{\rotatebox[origin=c]{-90}{\includegraphics[height=0.425\textwidth]{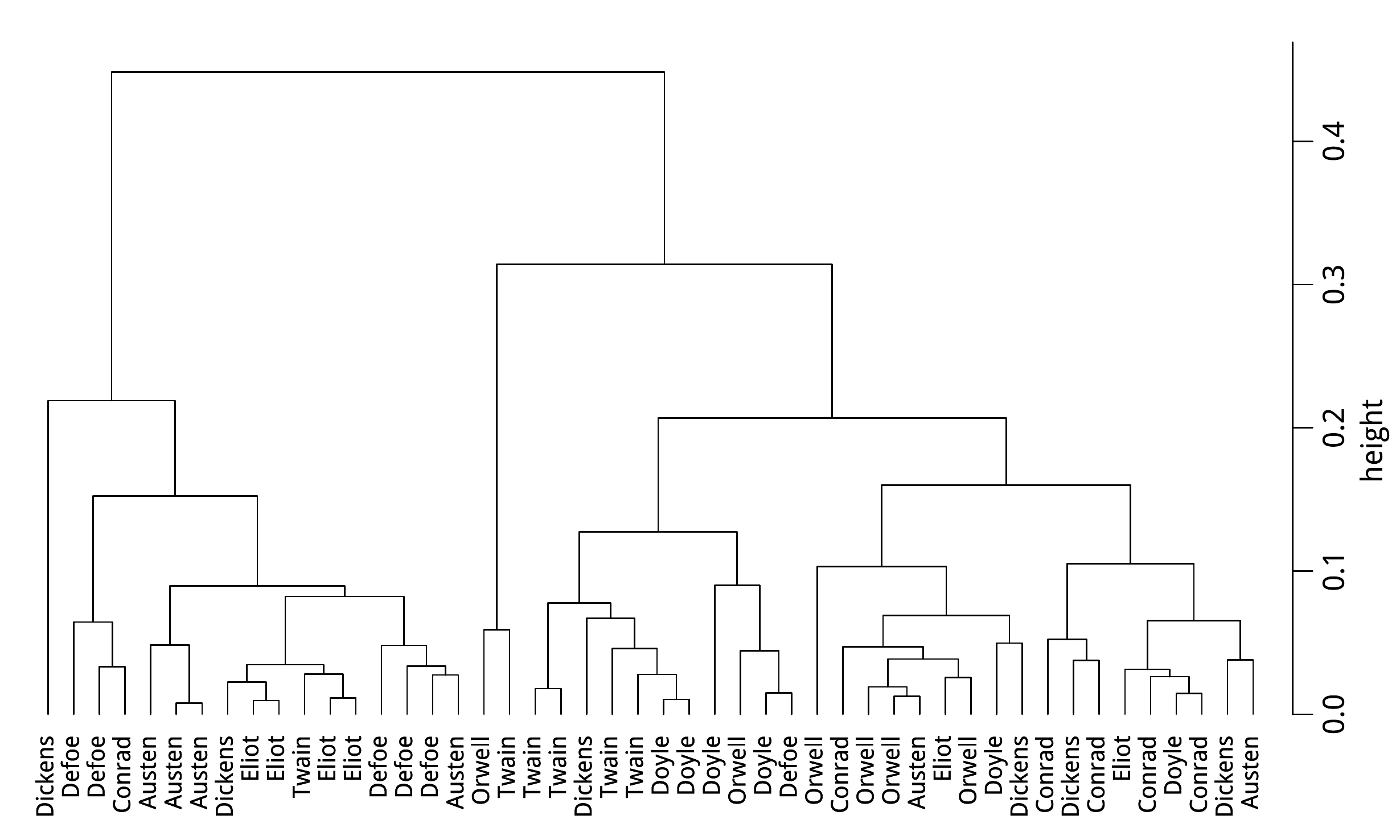}}}
\qquad
\subfigure[]{\rotatebox[origin=c]{-90}{\includegraphics[height=0.425\textwidth]{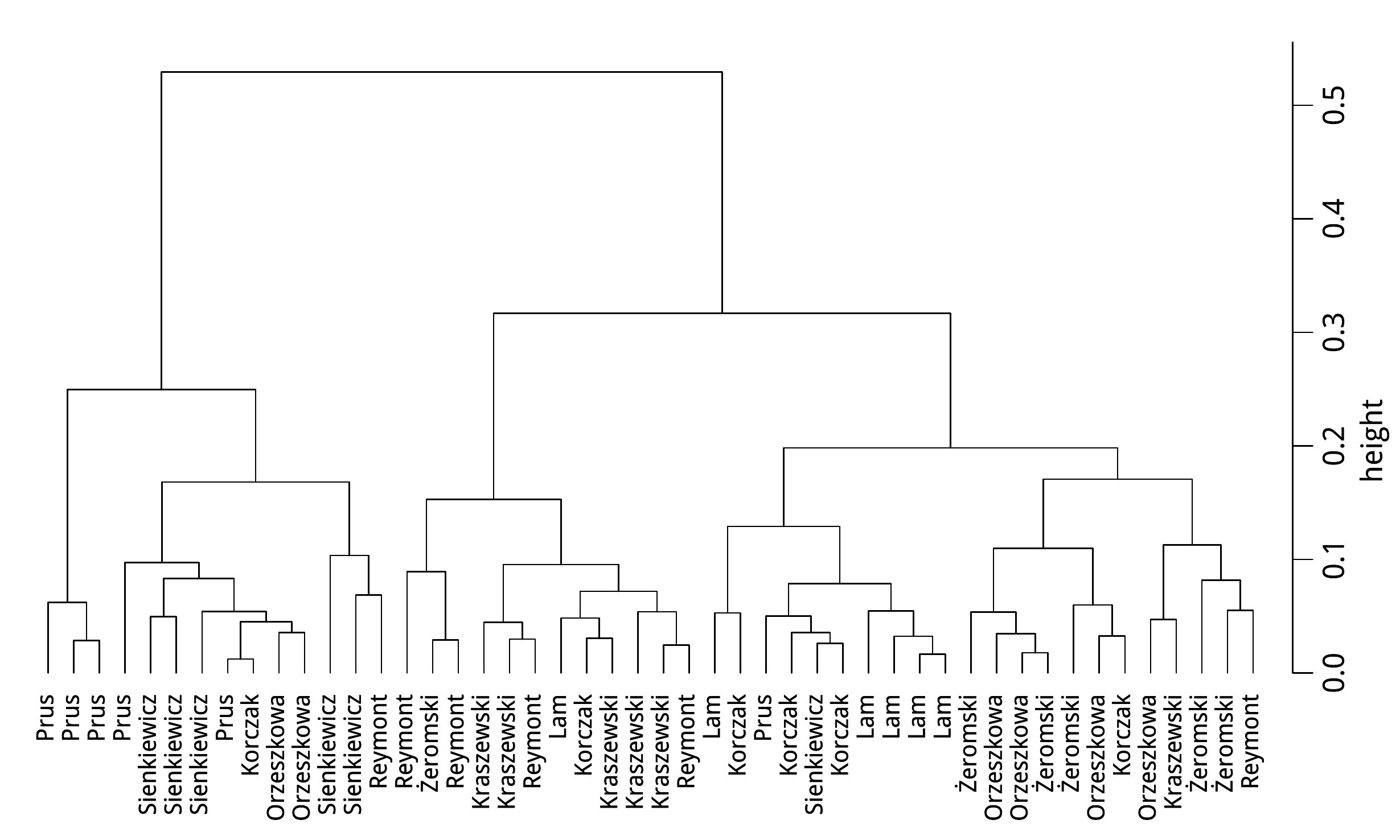}}}
\caption{The dendrograms of the hierarchical clustering of (a) English and (b) Polish books, in the space of \textbf{unweighted global characteristics of networks}. Each text is labeled by the surname of its author.}
\label{fig_dendro_indiv_unweighted}
\end{figure}

\begin{figure}[htbp]
\centering
\subfigure[]{\includegraphics[width=0.425\textwidth]{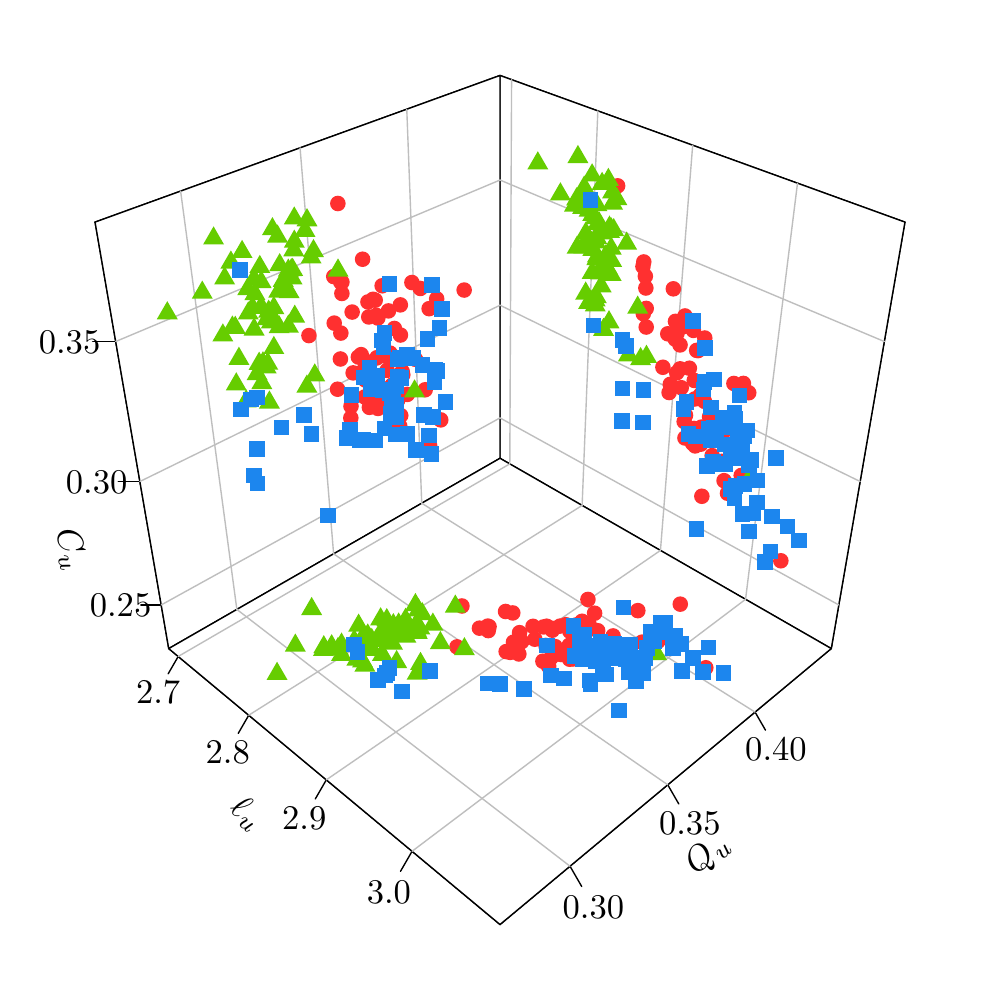}}
\subfigure[]{\includegraphics[width=0.425\textwidth]{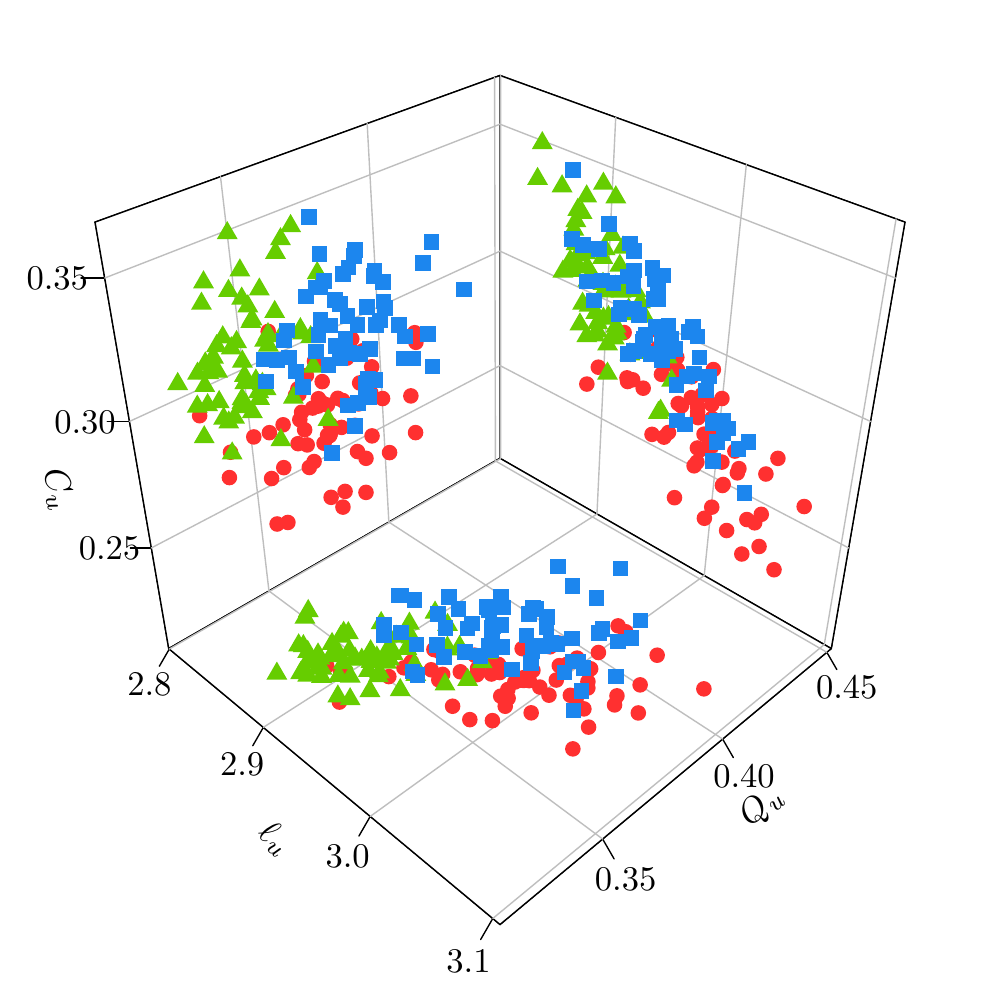}}

\subfigure[]{\includegraphics[width=0.425\textwidth]{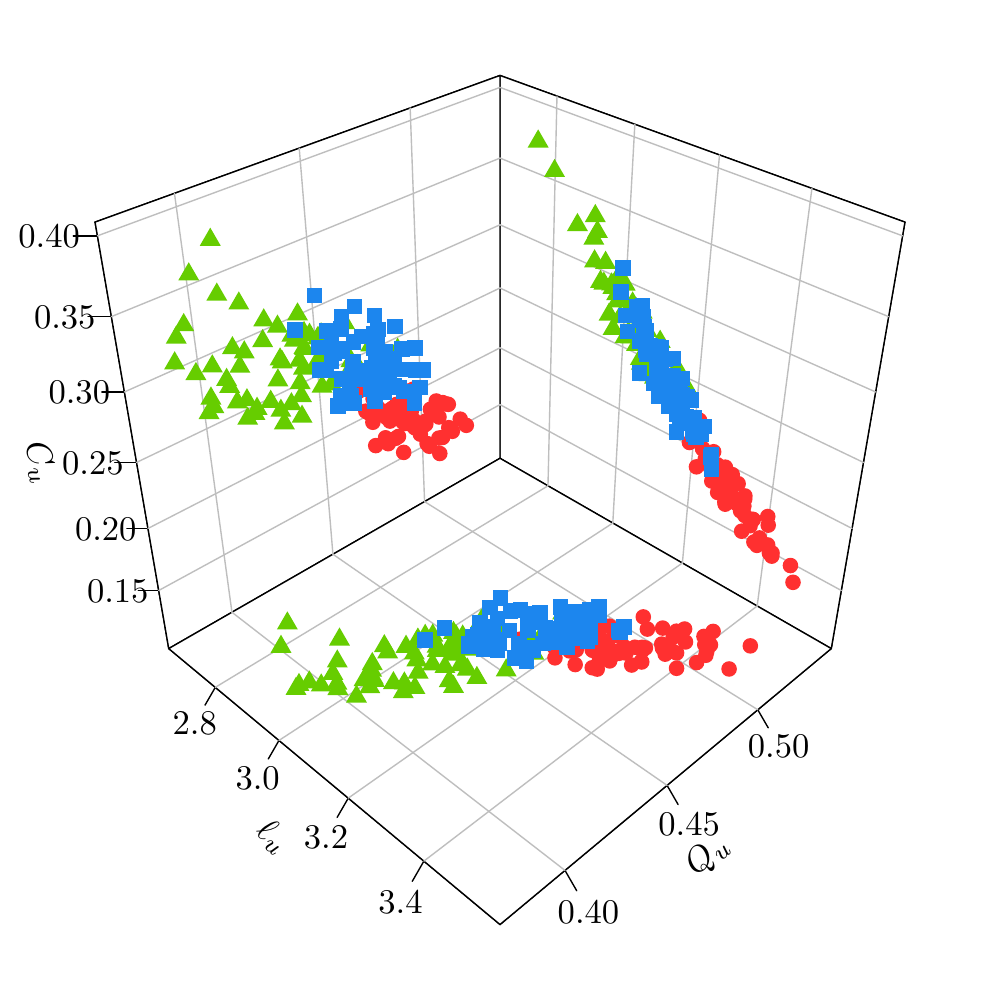}}
\subfigure[]{\includegraphics[width=0.425\textwidth]{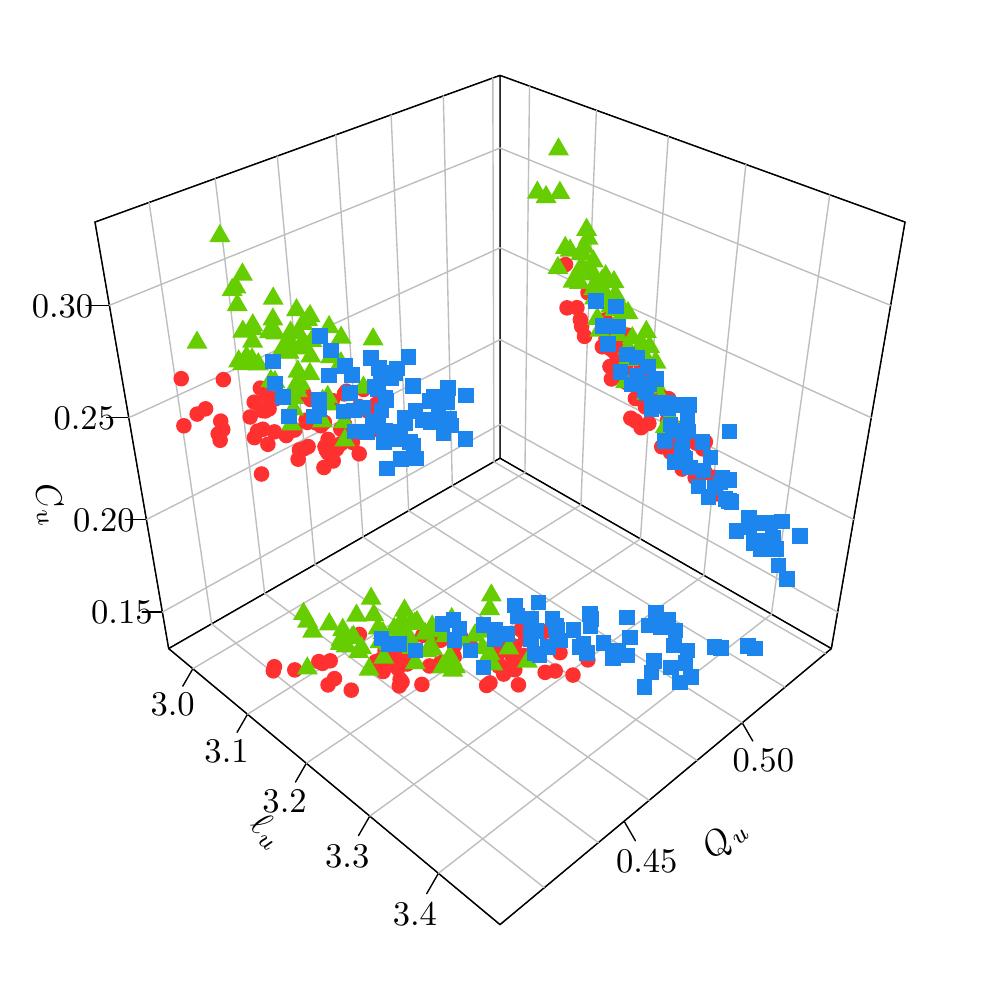}}
\caption{The projections of the triplets of network characteristics $\left(\ell_u,Q_u,C_u\right)$ onto planes $\left(\ell_u, Q_u\right)$, $\left(\ell_u, C_u\right)$, $\left(Q_u, C_u\right)$, for texts in English (a,~b) and Polish (c,~d). Each triplet of characteristics pertains to one chunk of text of length 5000 words. Texts samples were randomly chosen from all of the studied works of considered authors. Different markers denote different authors - red dots, green triangles and blue squares denote respectively: Charles Dickens, Daniel Defoe and Mark Twain in (a), George Eliot, Jane Austen, Joseph Conrad in (b), Władysław Reymont, Janusz Korczak, Jan Lam in (c) and Henryk Sienkiewicz, Józef Ignacy Kraszewski, Stefan Żeromski in (d). It can be seen that points representing texts of different authors tend to occupy different regions of space. Note that the characteristics are not normalized, as all the samples have the same length.}
\label{fig_scatterplots}
\end{figure}

The set of books is then divided according to the language, and hierarchical clustering is performed again, in the already separated sets of books in English and Polish. The outcome of this analysis (Figure \ref{fig_dendro_indiv_unweighted}) reveals the emergence of small clusters (consisting of 2-3 texts) corresponding to certain authors. This suggests the existence of connection between the authorship of the text and the structure of related network, expressed by appropriate quantities. Such hypothesis may also be supported by analysing the samples of text much shorter than the whole books. Figure \ref{fig_scatterplots} presents the plots of characteristics $\ell_u$, $Q_u$, $C_u$ of networks that are created from text samples randomly drawn from the books of selected authors, each sample containing 5000 words. It can be observed that although derived from different books, the networks corresponding to texts of the same author tend to be similar in terms of calculated characteristics. This effect is present for any 3-dimensional plot of an arbitrarily chosen triad of the studied characteristics $\ell_u$, $Q_u$, $C_u, r_u$.

To further examine the capability of network parameters of texts to distinguish between authors, statistical classification by an ensemble of decision trees is applied. The analysis is performed in the following way. For a given language, 4 randomly chosen books of each author are taken to form the training set, and the remaining 2 books contribute to the test set. Then the classifier (an ensemble of 100 decision trees) is trained on the training set to categorize books with respect to their author - each book constitutes one observation, whose attributes are the calculated network parameters. After that, the classifier categorizes observations in the test set. Picking training and test sets, training the classifier, and performing classification in the test set is repeated 10000 times. The average accuracy of classifier (the fraction of correct classifications) in the test set is treated as a measure of the possibility to distinguish between the authors. The reasoning behind this approach can be understood as follows: the greater the dissimilarities between individual authors (in terms of the network parameters), the easier it should be to train an accurate classifier and to categorize the observations in the test set correctly. Thus, the classifier's performance, which is given by the average accuracy in the test set, may serve as an indicator of network parameters' potential to distinguish between different writing styles. If the texts were not distinguishable at all, the classification would not be much different from a random choice, and in that case the average accuracy would be equal to $1/n$, where $n$ denotes the number of considered authors.

The detailed results of the classification are presented in Table \ref{tab_clas_global_only_unweighted}, which is organized as follows: a number in the $i$-th row and $j$-th column is the probability of classifying a text of the $i$-th author as a text of the $j$-th author, obtained by counting such classifications in the test set and dividing the number of counts by the number of performed repetitions of the test set selections (10000). The probabilities of correct classifications reside on the diagonal of the table. The sum of values in each row is equal to~1 - as it is the probability of assigning a text to any available author; the sums of values in columns are not given any constraint, as they do not have the interpretation of probability.

The obtained overall average accuracies of classifiers (the overall probabilities of correct classifications) in the test set are: 35\% with standard deviation of 10\% for English, and 41\% with standard deviation of 10\% for Polish. Although these values are higher than 12.5\% corresponding to a random choice, they are still too low to state that texts are clearly separated. Therefore it is reasonable to look for possible improvements. This is done by including the weighted versions of the network parameters in the calculation, in the next step of the analysis.

At this point, it is worth noting that the results obtained by the clustering and the classification are in decent accordance with each other. Comparing Figure \ref{fig_dendro_indiv_unweighted} and Table \ref{tab_clas_global_only_unweighted} reveals that the texts of the authors who are likely to be correctly classified are typically grouped together in clustering (for example, the texts of D. Defoe, J. Kraszewski, J. Lam), while the texts of authors weakly distinguished in classification are scattered over the dendrogram (C. Dickens, J. Korczak).

\begin{table}[htbp]
\small
\centering
\caption{The results of the classification of (a) English (b) Polish books with respect to the authorship, in the space of \textbf{unweighted global characteristics of networks}. A number in the $i$-th row and $j$-th column is the probability of classifying a text of $i$-th author as a text of $j$-th author. \label{tab_clas_global_only_unweighted}}
\vspace{0.2\baselineskip}
\subtable[The classification of English books. The authors are denoted by the two first letters of their surnames: Au -  Austen, Co - Conrad, De - Defoe, Di - Dickens, Do - Doyle, El - Eliot, Or - Orwell, Tw - Twain.]
{
\setlength{\tabcolsep}{4pt}
\setlength{\extrarowheight}{0.25\tabcolsep}

\begin{tabular}{c|cccccccc}
   & Au & Co & De & Di & Do & El & Or & Tw \\ 
   \hline
Au & \textbf{.33} & .22 & .17 & .06 & .01 & .11 & .10 & .00 \\ 
  Co & .08 & \textbf{.34} & .01 & .21 & .02 & .11 & .14 & .09 \\ 
  De & .21 & .00 & \textbf{.54} & .00 & .09 & .04 & .12 & .00 \\ 
  Di & .18 & .21 & .00 & \textbf{.05} & .23 & .18 & .08 & .07 \\ 
  Do & .00 & .12 & .06 & .09 & \textbf{.28} & .03 & .24 & .18 \\ 
  El & .10 & .22 & .16 & .05 & .04 & \textbf{.39} & .04 & .00 \\ 
  Or & .10 & .04 & .09 & .07 & .23 & .12 & \textbf{.18} & .17 \\ 
  Tw & .00 & .00 & .00 & .03 & .12 & .16 & .03 & \textbf{.66} \\ 
  \end{tabular}

}
\hfill
\subtable[The classification of Polish books. The authors are denoted by the two first letters of their surnames: Ko - Korczak, Kr - Kraszewski, La - Lam, Or - Orzeszkowa, Pr - Prus, Re - Reymont, Si - Sienkiewicz, Że - Żeromski.]
{
\setlength{\tabcolsep}{4pt}
\setlength{\extrarowheight}{0.25\tabcolsep}

\begin{tabular}{c|cccccccc}
   & Ko & Kr & La & Or & Pr & Re & Si & Że \\ 
   \hline
Ko & \textbf{.18} & .23 & .14 & .18 & .04 & .00 & .09 & .14 \\ 
  Kr & .16 & \textbf{.57} & .00 & .02 & .00 & .18 & .03 & .04 \\ 
  La & .21 & .06 & \textbf{.54} & .02 & .05 & .00 & .08 & .04 \\ 
  Or & .14 & .00 & .02 & \textbf{.26} & .11 & .00 & .22 & .25 \\ 
  Pr & .02 & .00 & .09 & .12 & \textbf{.65} & .00 & .10 & .02 \\ 
  Re & .05 & .21 & .00 & .01 & .00 & \textbf{.44} & .07 & .22 \\ 
  Si & .04 & .00 & .18 & .21 & .09 & .00 & \textbf{.46} & .02 \\ 
  Że & .19 & .10 & .01 & .32 & .00 & .17 & .00 & \textbf{.21} \\ 
  \end{tabular}

}
\end{table}

\subsection{Including weighted characteristics}

The analysis of weighted network parameters is performed in the same manner as in the case of the unweighted ones. The whole procedure described above is repeated, but this time the networks created from texts are weighted networks, and the characteristics (average shortest path length, clustering coefficient, assortativity coefficient, modularity) are calculated in both the unweighted and the weighted version. The vectors of parameters calculated for each book (this time they are 8-dimensional) again serve as an input of data clustering and classification algorithms.

Table \ref{tab_clas_global_weighted_and_unweighted} presents the outcome of the classification. It can be seen that including weighted network characteristics into analysis gives a slight improvement of the distinguishability between the English-language authors - the overall average accuracy of English texts classification is 42\% with standard deviation of 11\%. For Polish books, the increase in the classifier's performance is rather negligible, the accuracy obtained is 44\% with standard deviation of 11\%. For some individual authors, the probability of correct classification decreases after including the weighted characteristics of networks. The results of hierarchical clustering in the space with 4 additional dimensions do not change significantly - they are analogous to what can be seen in Figure \ref{fig_dendro_indiv_unweighted}; for this reason the dendrograms are not presented here.

The classification of texts using only weighted characteristics $\widetilde{\ell_w}$, $\widetilde{Q_w}$, $\widetilde{C_w}, \widetilde{r_w}$ gives results very similar to these from the analysis of the unweighted networks - the classification accuracy for English and Polish is 37\% and 41\% with standard deviations of 11\% and 10\%, respectively. This similarity along with the fact that the classification for both the unweighted and weighted parameters together is not much more accurate, suggests that the information carried by these two types of characteristics is to a large extent overlapping.

\begin{table}[htbp]
\small
\centering
\caption{The results of the classification of (a) English (b) Polish books with respect to the authorship, in the space of \textbf{unweighted and weighted global characteristics of networks}. A number in the $i$-th row and $j$-th column is the probability of classifying a text of $i$-th author as a text of $j$-th author.
\label{tab_clas_global_weighted_and_unweighted}}
\vspace{0.2\baselineskip}
\subtable[The classification of English books. The authors are denoted by the two first letters of their surnames: Au -  Austen, Co - Conrad, De - Defoe, Di - Dickens, Do - Doyle, El - Eliot, Or - Orwell, Tw - Twain.]
{
\setlength{\tabcolsep}{4pt}
\setlength{\extrarowheight}{0.25\tabcolsep}

\begin{tabular}{c|cccccccc}
   & Au & Co & De & Di & Do & El & Or & Tw \\ 
   \hline
Au & \textbf{.20} & .17 & .18 & .12 & .00 & .22 & .11 & .00 \\ 
  Co & .12 & \textbf{.62} & .00 & .06 & .02 & .07 & .03 & .08 \\ 
  De & .11 & .00 & \textbf{.52} & .11 & .09 & .14 & .01 & .02 \\ 
  Di & .16 & .03 & .03 & \textbf{.21} & .22 & .18 & .09 & .08 \\ 
  Do & .02 & .10 & .00 & .22 & \textbf{.33} & .01 & .14 & .18 \\ 
  El & .17 & .16 & .13 & .04 & .00 & \textbf{.46} & .00 & .04 \\ 
  Or & .06 & .03 & .05 & .04 & .05 & .01 & \textbf{.55} & .21 \\ 
  Tw & .00 & .01 & .03 & .02 & .10 & .13 & .22 & \textbf{.49} \\ 
  \end{tabular}

}
\hfill
\subtable[The classification of Polish books. The authors are denoted by the two first letters of their surnames: Ko - Korczak, Kr - Kraszewski, La - Lam, Or - Orzeszkowa, Pr - Prus, Re - Reymont, Si - Sienkiewicz, Że - Żeromski.]
{
\setlength{\tabcolsep}{4pt}
\setlength{\extrarowheight}{0.25\tabcolsep}

\begin{tabular}{c|cccccccc}
   & Ko & Kr & La & Or & Pr & Re & Si & Że \\ 
   \hline
Ko & \textbf{.22} & .22 & .16 & .09 & .01 & .01 & .15 & .14 \\ 
  Kr & .21 & \textbf{.35} & .03 & .07 & .01 & .29 & .00 & .04 \\ 
  La & .06 & .04 & \textbf{.55} & .05 & .12 & .00 & .03 & .15 \\ 
  Or & .16 & .04 & .02 & \textbf{.34} & .13 & .00 & .12 & .19 \\ 
  Pr & .01 & .00 & .08 & .18 & \textbf{.63} & .00 & .10 & .00 \\ 
  Re & .01 & .12 & .00 & .00 & .00 & \textbf{.50} & .18 & .19 \\ 
  Si & .10 & .00 & .07 & .06 & .15 & .01 & \textbf{.61} & .00 \\ 
  Że & .22 & .03 & .08 & .08 & .00 & .28 & .00 & \textbf{.31} \\ 
  \end{tabular}

}
\end{table}


\subsection{Local characteristics}

The analysis of local parameters of networks is an approach different from what has been done in previous steps. Instead of studying the properties of texts by determining the quantities describing the structure of the whole networks, here the characteristics of the individual vertices (words) are considered. In each language, $n$ most frequently occurring words are chosen from the whole set of texts. Each book is transformed into a network, and the characteristics of vertices corresponding to selected words are determined. The characteristics taken into account are: vertex degree, local clustering coefficient, and average shortest path length, all in both the unweighted and weighted versions. All of these quantities are normalized, in the same manner as in previous calculations - by dividing their value by the average value obtained for the same word in a randomized network. Vertex strength (weighted degree) is an exception here - since it is equal to the frequency of the related word multiplied by 2 (and frequency is invariant under text randomization), its normalized value should be equal to 1 for any word. Therefore $\widetilde{\text{str}}(v)$, the normalized strength of a vertex $v$, is defined as the strength of vertex $v$ in the original network, divided by the sum of strengths of all vertices in the same network: $\widetilde{\text{str}}(v) = \text{str}(v)/ \sum_{u \in V}{\text{str}(u)}$. This is equal to the word's relative frequency (the fraction of the total number of words in the text that comes from the occurrences of the considered word).

\begin{figure}[htb]
\centering
\subfigure[]{\rotatebox[origin=c]{-90}{\includegraphics[height=0.425\textwidth]{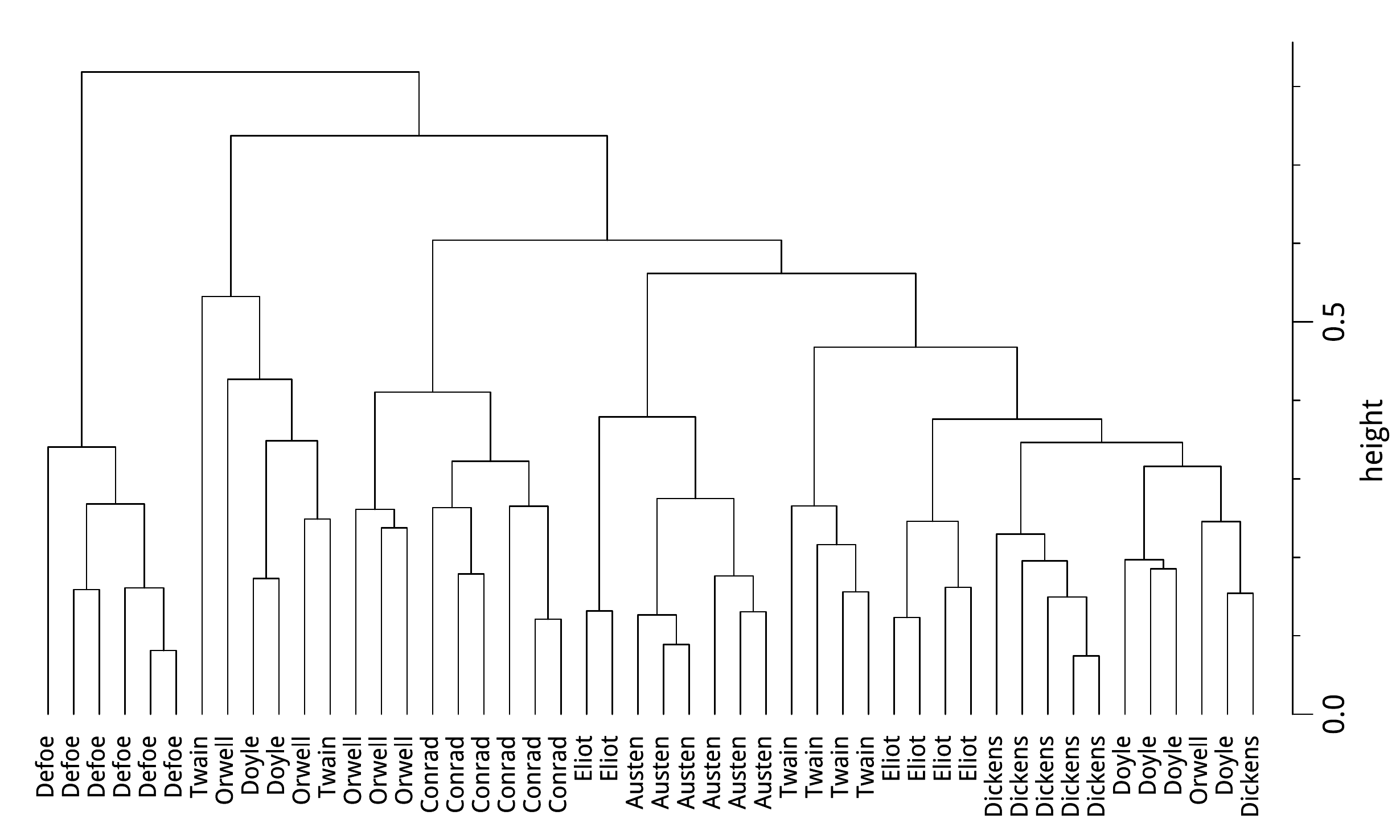}}}
\qquad
\subfigure[]{\rotatebox[origin=c]{-90}{\includegraphics[height=0.425\textwidth]{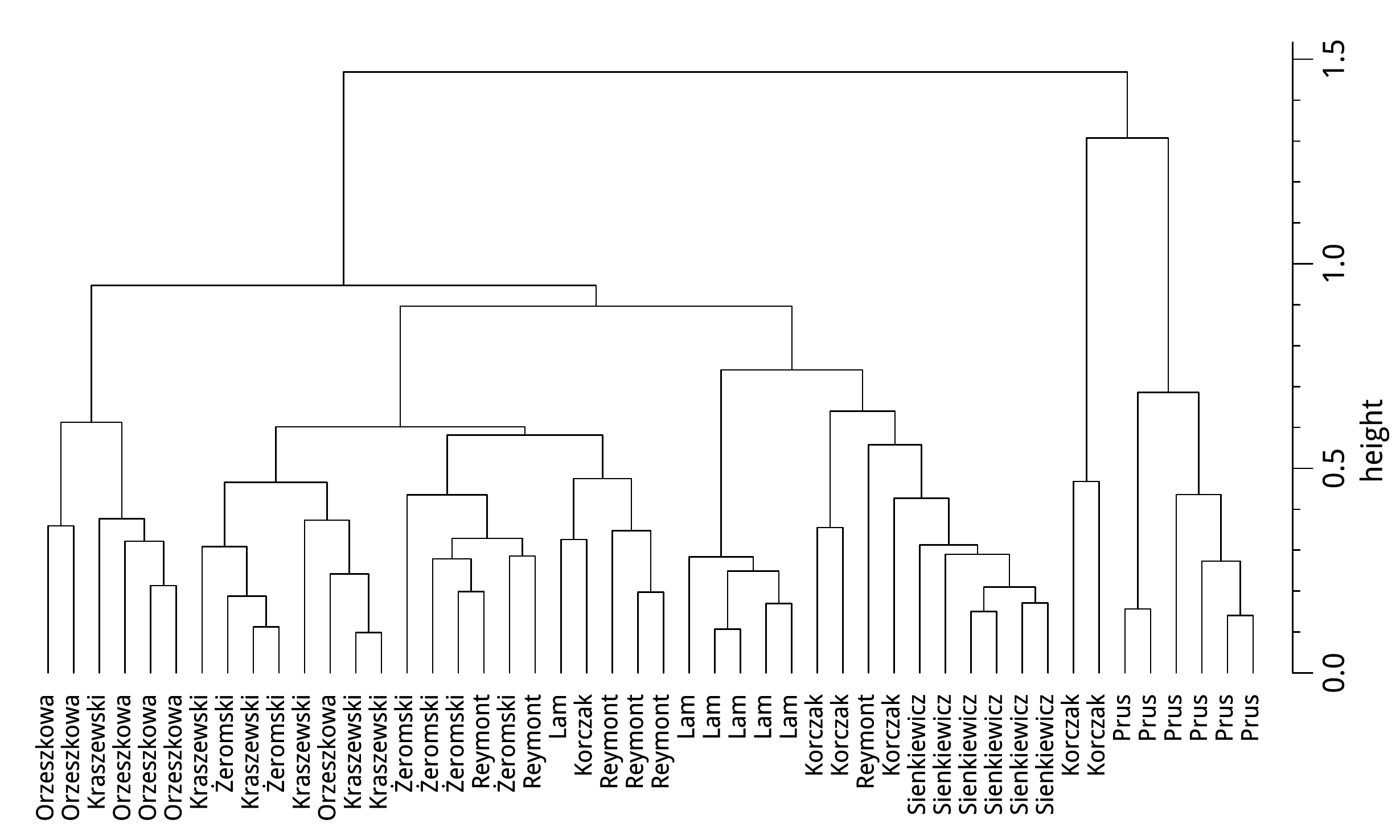}}}
\caption{The dendrograms of the hierarchical clustering of (a) English and (b) Polish books, in the space of the \textbf{weighted clustering coefficients of networks' nodes, corresponding to 12 most frequent words}. Each text is labeled by the surname of its author.}
\label{fig_dendro_local}
\end{figure}

The sets of parameters related to the $n$ most frequent words in each text are again supplied to the data clustering and classification algorithms. Figure \ref{fig_dendro_local} presents the dendrogram of the clustering in the 12-dimensional space of the weighted clustering coefficients of $n=12$ most frequent words. The results of text classification in this space are shown in Table \ref{tab_clas_local}. The obtained overall classification accuracy is 90\% with standard deviation of 8\% for English texts, and 86\% with standard deviation of 8\% for the Polish ones.

\begin{figure}[htbp]
\centering
\subfigure[]{{\includegraphics[width=0.49\textwidth]{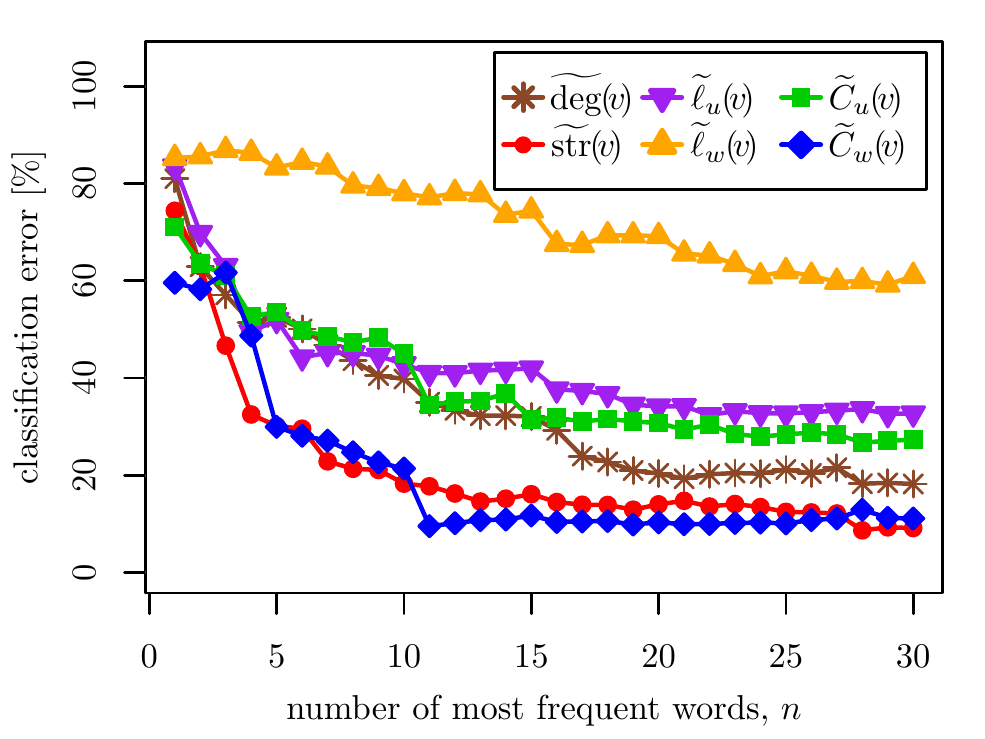}}}
\hfill
\subfigure[]{{\includegraphics[width=0.49\textwidth]{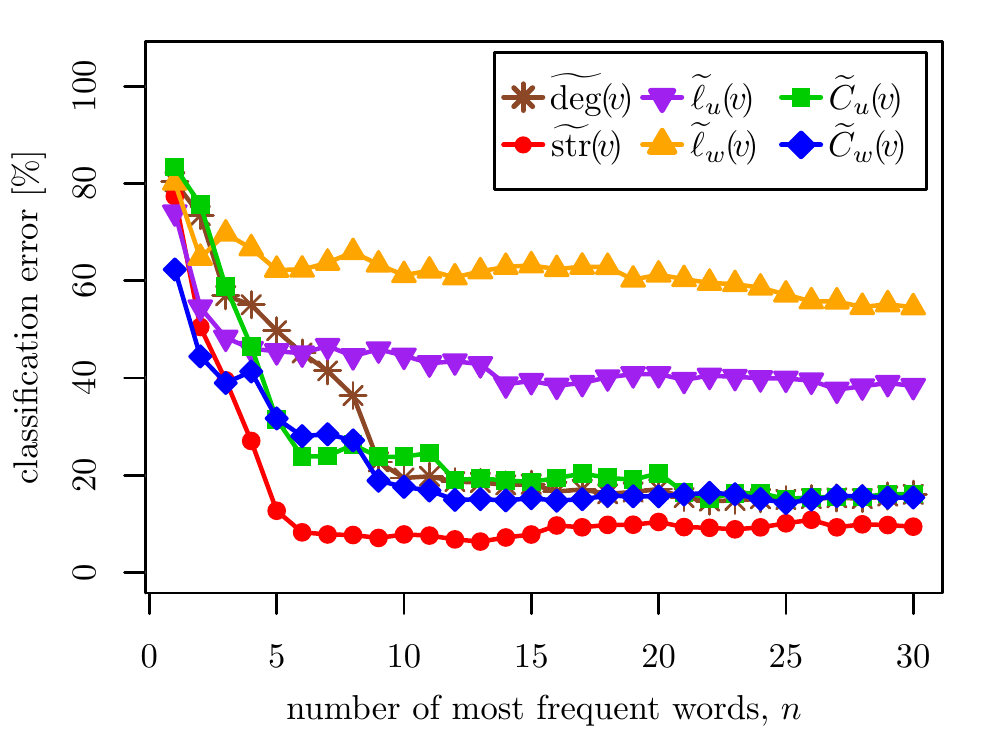}}}
\caption{The classification of books in the feature spaces constructed from a single network parameter determined for a set of $n$ words occurring most frequently in the whole collection of books. Charts (a) and (b) present the average classification error as a function of $n$, for English and Polish books, respectively. Each point on a chart represents the average classification error in the test set, obtained in one experiment. One experiment consists of selecting the $n$ most frequent words, calculating one network parameter for each of these words in each text, and performing the cross-validation of classification in the so obtained $n$-dimensional space, 10000 times. The studied network characteristics are (all normalized): vertex degree and strength ($\widetilde{\deg}(v)$, $\widetilde{\text{str}}(v)$), unweighted and weighted average shortest path length ($\widetilde{\ell}_u(v)$, $\widetilde{\ell}_w(v)$), unweighted and weighted clustering coefficient ($\widetilde{C}_u(v)$, $\widetilde{C}_w(v)$).}
\label{fig_accuracy_1_char}
\end{figure}

The choice of the number of words, $n=12$, and the weighted clustering coefficient $\widetilde{C}_w$ as the only parameter to consider in the above example is justified by studying the effectiveness of classification with each of the network characteristics separately. It can be noticed (Figure \ref{fig_accuracy_1_char}) that the weighted clustering coefficient gives the best results in English and one of the best in Polish for a wide range of the most frequent words studied. In both languages, it is sufficient to analyse the 11-12 most frequent words to obtain the accuracy of 85-90\%; further increase in the number of words does not improve the classifier's performance.

\begin{table}[htb]
\small
\centering
\caption{The results of the classification of (a) English (b) Polish books with respect to the authorship, in the space of the \textbf{weighted clustering coefficient of networks' nodes, corresponding to 12 most frequent words}. A number in the $i$-th row and $j$-th column is the probability of classifying a text of $i$-th author as a text of $j$-th author.
\label{tab_clas_local}}
\vspace{0.2\baselineskip}
\subtable[The classification of English books. The authors are denoted by the two first letters of their surnames: Au -  Austen, Co - Conrad, De - Defoe, Di - Dickens, Do - Doyle, El - Eliot, Or - Orwell, Tw - Twain.]
{
\setlength{\tabcolsep}{4pt}
\setlength{\extrarowheight}{0.25\tabcolsep}

\begin{tabular}{c|cccccccc}
   & Au & Co & De & Di & Do & El & Or & Tw \\ 
   \hline
Au & \textbf{.96} & .00 & .00 & .01 & .01 & .02 & .00 & .00 \\ 
  Co & .00 & \textbf{.92} & .00 & .00 & .00 & .04 & .04 & .00 \\ 
  De & .00 & .00 & \textbf{1.0} & .00 & .00 & .00 & .00 & .00 \\ 
  Di & .01 & .00 & .00 & \textbf{.88} & .04 & .00 & .00 & .07 \\ 
  Do & .01 & .00 & .00 & .05 & \textbf{.93} & .01 & .00 & .00 \\ 
  El & .02 & .02 & .00 & .03 & .01 & \textbf{.87} & .02 & .03 \\ 
  Or & .00 & .13 & .00 & .01 & .01 & .07 & \textbf{.78} & .00 \\ 
  Tw & .00 & .01 & .00 & .00 & .01 & .02 & .07 & \textbf{.89} \\ 
  \end{tabular}

}
\hfill
\subtable[The classification of Polish books. The authors are denoted by the two first letters of their surnames: Ko - Korczak, Kr - Kraszewski, La - Lam, Or - Orzeszkowa, Pr - Prus, Re - Reymont, Si - Sienkiewicz, Że - Żeromski.]
{
\setlength{\tabcolsep}{4pt}
\setlength{\extrarowheight}{0.25\tabcolsep}

\begin{tabular}{c|cccccccc}
   & Ko & Kr & La & Or & Pr & Re & Si & Że \\ 
   \hline
Ko & \textbf{.73} & .00 & .00 & .00 & .00 & .06 & .03 & .18 \\ 
  Kr & .00 & \textbf{.80} & .00 & .12 & .00 & .00 & .05 & .03 \\ 
  La & .00 & .00 & \textbf{1.0} & .00 & .00 & .00 & .00 & .00 \\ 
  Or & .00 & .15 & .00 & \textbf{.83} & .00 & .00 & .01 & .01 \\ 
  Pr & .00 & .00 & .00 & .00 & \textbf{1.0} & .00 & .00 & .00 \\ 
  Re & .04 & .00 & .00 & .00 & .00 & \textbf{.77} & .02 & .17 \\ 
  Si & .00 & .05 & .00 & .00 & .04 & .01 & \textbf{.90} & .00 \\ 
  Że & .01 & .15 & .00 & .00 & .00 & .03 & .02 & \textbf{.79} \\ 
  \end{tabular}

}
\end{table}

From the results obtained, it can be obviously concluded that analysing the local parameters of selected words leads to much better distinguishability between the authors than studying the global characteristics of the networks. In the "local approach" the number of considered attributes of each text can be much greater than when studying the networks globally, as one can choose the number of analysed words arbitrary. However, even for equal dimension of the attribute space, the networks' local properties (weighted clustering coefficient, for example) provide the significantly higher texts' classification accuracy.

It can be found in the literature on computational stylometry that a very often utilized approach to authorship attribution is representing a text as a collection of words along with their frequencies (the so-called \textit{bag-of-words} representation), and comparing particular words' frequencies is among the most reliable methods to discriminate between authors \cite{Vel2001, Madigan05, Zheng2006, Stamatatos2009, Koppel2009, Neal2017}. As mentioned in Section \ref{sec_data}, the normalized node strength in a weighted word-adjacency network is equal to the corresponding word's relative frequency; therefore network analysis may be viewed as a generalization of the bag-of-words approach: it incorporates all the information about the numbers of word occurrences. 

\begin{figure}[htbp]
\centering
\subfigure[]{{\includegraphics[width=0.49\textwidth]{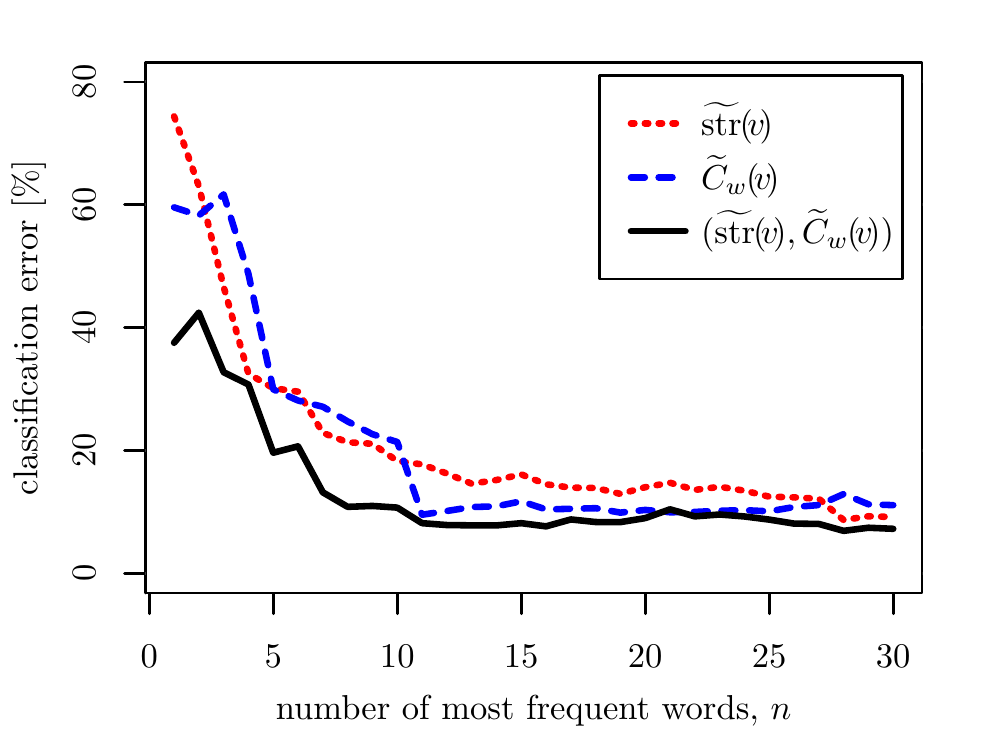}}}
\hfill
\subfigure[]{{\includegraphics[width=0.49\textwidth]{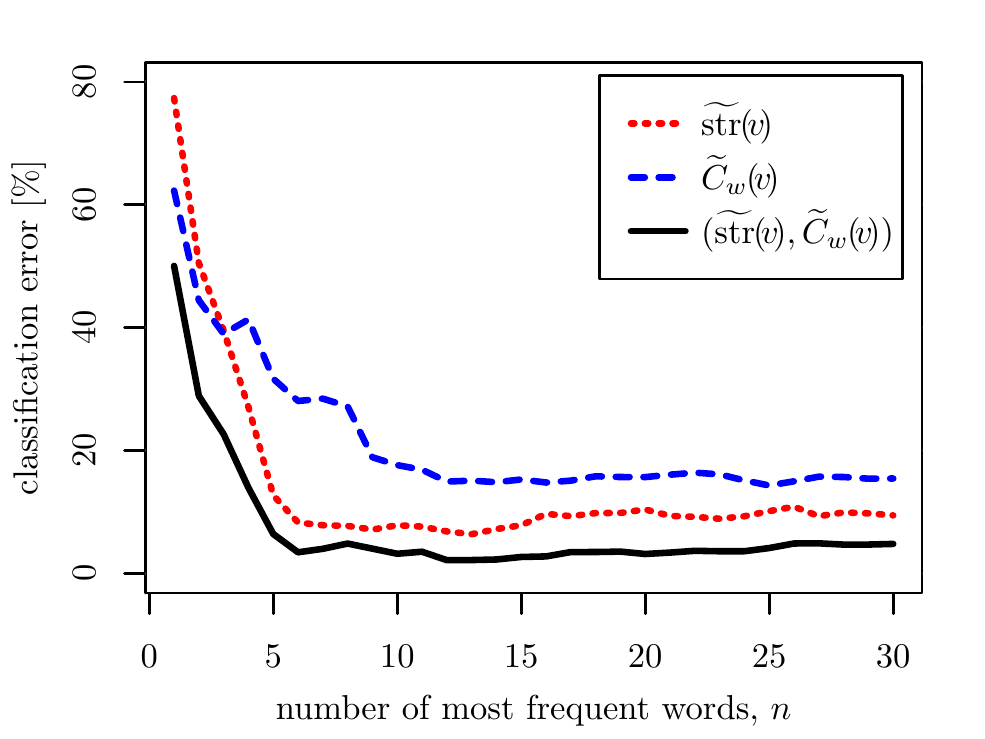}}}
\caption{The classification of books in the feature spaces constructed from the sets of network parameters, determined for a set of $n$ words occurring most frequently in the whole collection of books. Charts (a) and (b) present the average (obtained from 10000 repetitions of cross-validation) classification error in the test set, as a function of $n$, for English and Polish books, respectively. 3 sets of quantities describing words in texts were investigated, namely: (1) normalized vertex strength $\widetilde{\text{str}}(v)$, (2) normalized weighted clustering coefficient $\widetilde{C}_w(v)$, (3) normalized vertex strength $\widetilde{\text{str}}(v)$ together with normalized weighted clustering coefficient $\widetilde{C}_w(v)$.}
\label{fig_accuracy_freq_clustering}
\end{figure}

The fact that the word frequency works well in identifying authors can be observed in Figure \ref{fig_accuracy_1_char}. The classification based on normalized vertex strength leads to the best or one of the best results among all the studied network characteristics. A question arises, whether it is viable to introduce the network formalism when a satisfying result can be obtained by studying only the numbers of word occurrences. It turns out that a profit can be made from combining the purely network-based parameters with the frequencies. For example, a classification in the space constructed from both the relative frequencies and the normalized weighted clustering coefficients leads to accuracy higher than achievable by any of the two methods separately, of course at the cost of doubling the dimension of attribute space. This is shown in Figure \ref{fig_accuracy_freq_clustering}. Furthermore, these two parameters seem to exploit all the available information that is useful with respect to authorship recognition, because including further network characteristics into the classification does not improve its accuracy.

From a practical point of view, the above result may be useful in classifying texts in which the set of available words is limited. This may happen, for example, when studied texts are much shorter than these considered here and the words that are the most frequent there are topic-specific and do not carry valuable information about an individual writing style. According to the Zipf's law, the frequency of the $n$-th most frequent word in a text is proportional to $1/n$, so the number of words with the number of occurrences high enough to be reasonably included into analysis decreases substantially when texts are shorter.

The results presented in Figure \ref{fig_accuracy_1_char} may suggest which properties of the networks seem to be in a way universal for the language, and which differ among authors. Aside from word frequency, the weighted clustering coefficient turned out to be most effective in capturing the diversity of writing styles, when considering both the English and Polish texts. Clustering coefficient $C_w(v)$ (unnormalized) of a vertex $v$ describes the structure of the neighbourhood of $v$ - it measures the contribution to the strength of $v$ coming from pairs of its neighbours that are connected by an edge. The normalized coefficient, $\widetilde{C}_w(v)$, reflects the deviation of this structure from the structure that would be typically observed in a randomized network. The fact that it allows for a quite accurate classification suggests that the organization of such local structures in the word-adjacency networks may be considered as a feature of individual language that is able to distinguish one author from another. 

\begin{figure}[htbp]
\centering
\subfigure[]{{\includegraphics[width=0.49\textwidth]{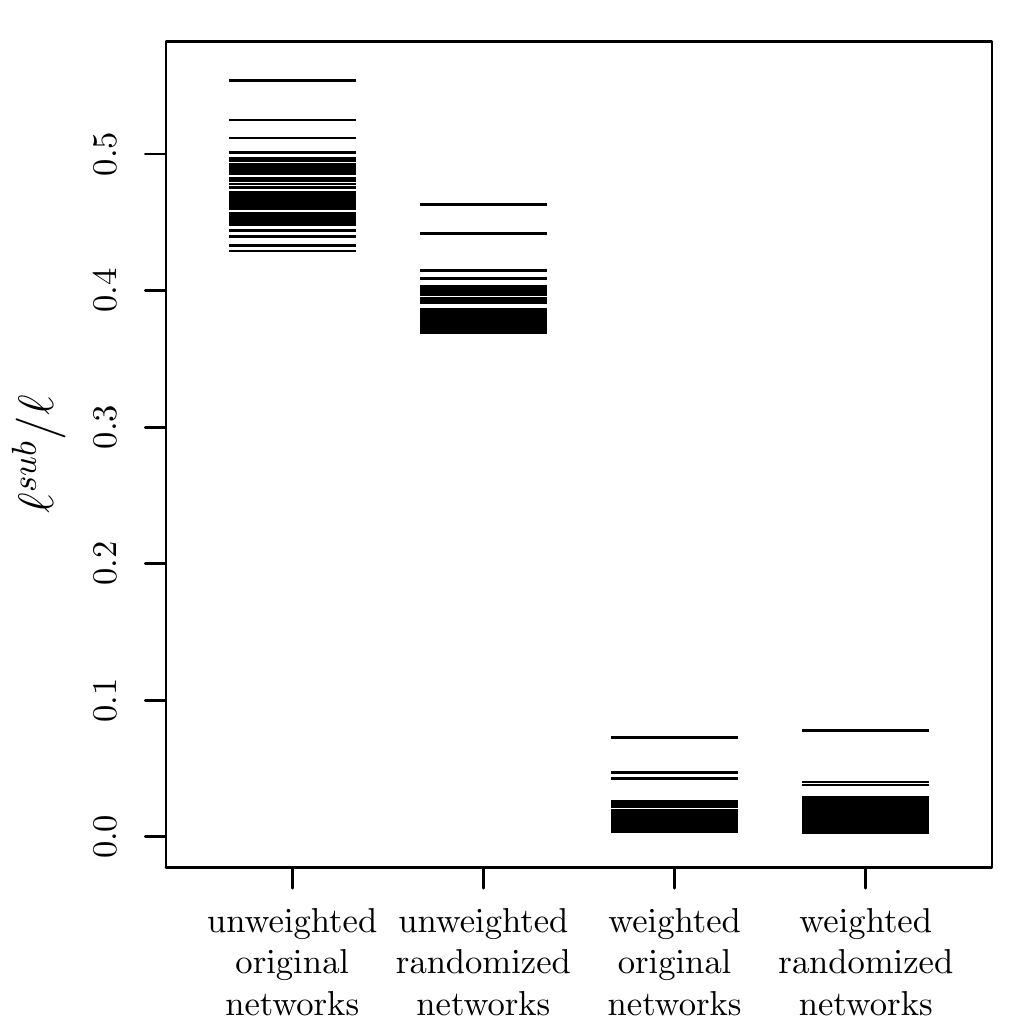}}}
\hfill
\subfigure[]{{\includegraphics[width=0.49\textwidth]{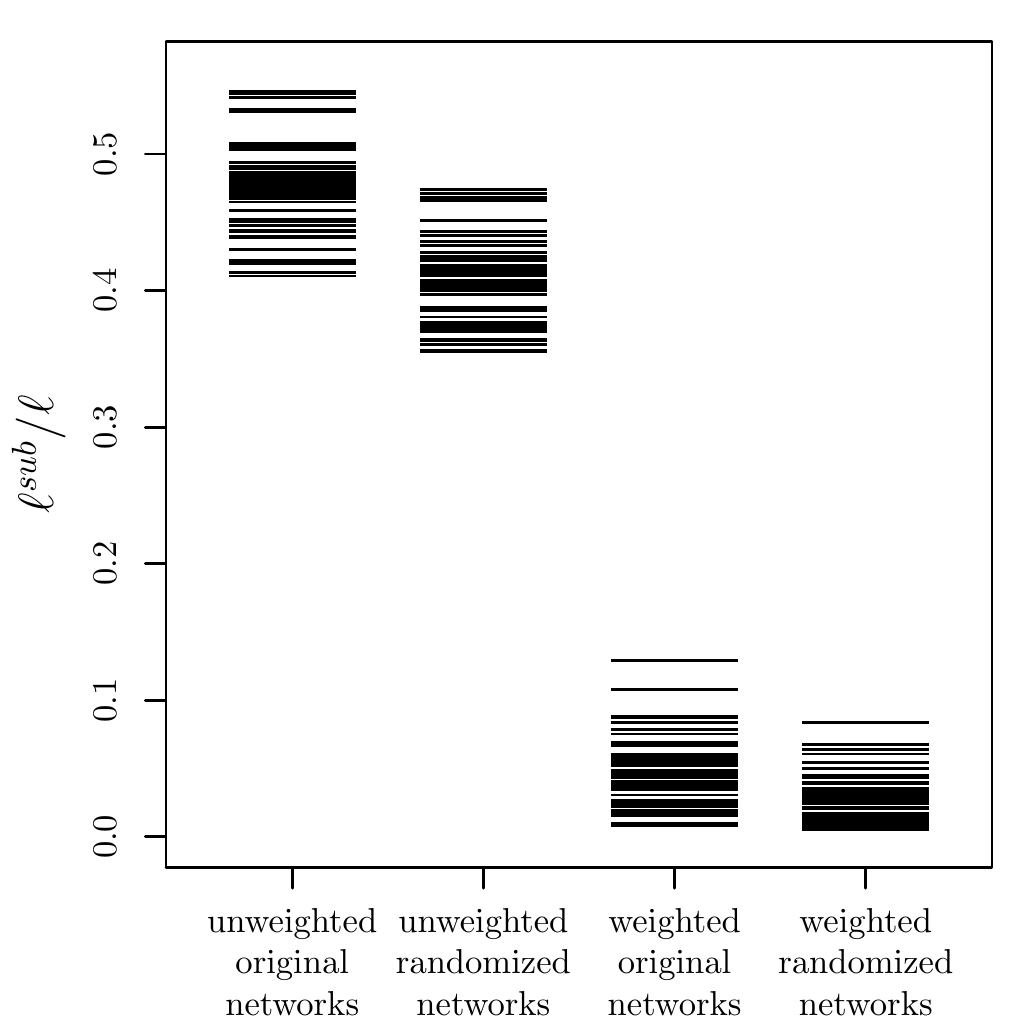}}}
\caption{The distributions of the quantity $\ell^{sub}/\ell$, for networks constructed from (a) English and (b) Polish books.  $\ell^{sub}$ denotes the global average shortest path length in a subnetwork consisting of 50 most frequent words, and $\ell$ denotes the average shortest path length in the whole network. Both original and randomized networks are considered, in both weighted and unweighted version.}
\label{fig_paths_subgraphs}
\end{figure}

On the other hand, the average shortest path length, especially in its weighted version, gives the weakest classifier. Thus, one may anticipate that this quantity is shared among the word-adjacency networks even when they are associated with different language users. Indeed, during the construction of a word-adjacency network from any sufficiently long text, a set of the most frequent words form a densely-connected cluster. Such cluster is a subnetwork, whose any two vertices are connected by a path of low length. This can be noticed when comparing the average shortest path length in such a subnetwork, $\ell^{sub}$, with the average shortest path length in the whole network, $\ell$. Figure \ref{fig_paths_subgraphs} presents the strip charts of the distributions of the quotient $\ell^{sub}/\ell$, for all the studied networks, with the subnetworks consisting of the 50 most frequent words. For the unweighted networks, the value of $\ell^{sub}/\ell$ is usually around 0.4-0.5 (because the smallest possible distance between two distinct vertices is equal to 1); however, in weighted networks it is typically smaller than 0.1 (as distances can be made arbitrarily small when edges' weight increase). Hence, in the weighted networks, for any $v$ being one of, say, the 50 most frequent words, the local average shortest path length $\ell(v)$ has roughly the same value. This is the reason why the average shortest path length gives a poor insight into the way of using particular words by different authors. The existence of a cluster with high edge density, composed of the most frequent words, is observable both in original and randomized networks; therefore it can be viewed as an effect of the principle of word-adjacency network construction, related to the word frequency distribution.

It may be interesting how including punctuation into the calculations affects the distinguishability of authors. It turns out that the utilized approach of treating punctuation marks as usual words \cite{Kulig2017} is justified: the classification accuracy decreases if punctuation is removed, even when the dimension of the attribute space is preserved (the removed punctuation marks are replaced with subsequent words from the list of the most frequent words). For example, for the normalized weighted clustering coefficient, $\widetilde{C}_w(v)$, of $n=12$ the most frequent words (not including punctuation marks), the average classification accuracy is 75\% for English and 76\% for Polish. A substantial drop from the values obtained previously suggests that the role of punctuation in expressing individual writing style is non-negligible.

\begin{figure}[htbp]
\centering
\includegraphics[width=0.7\textwidth]{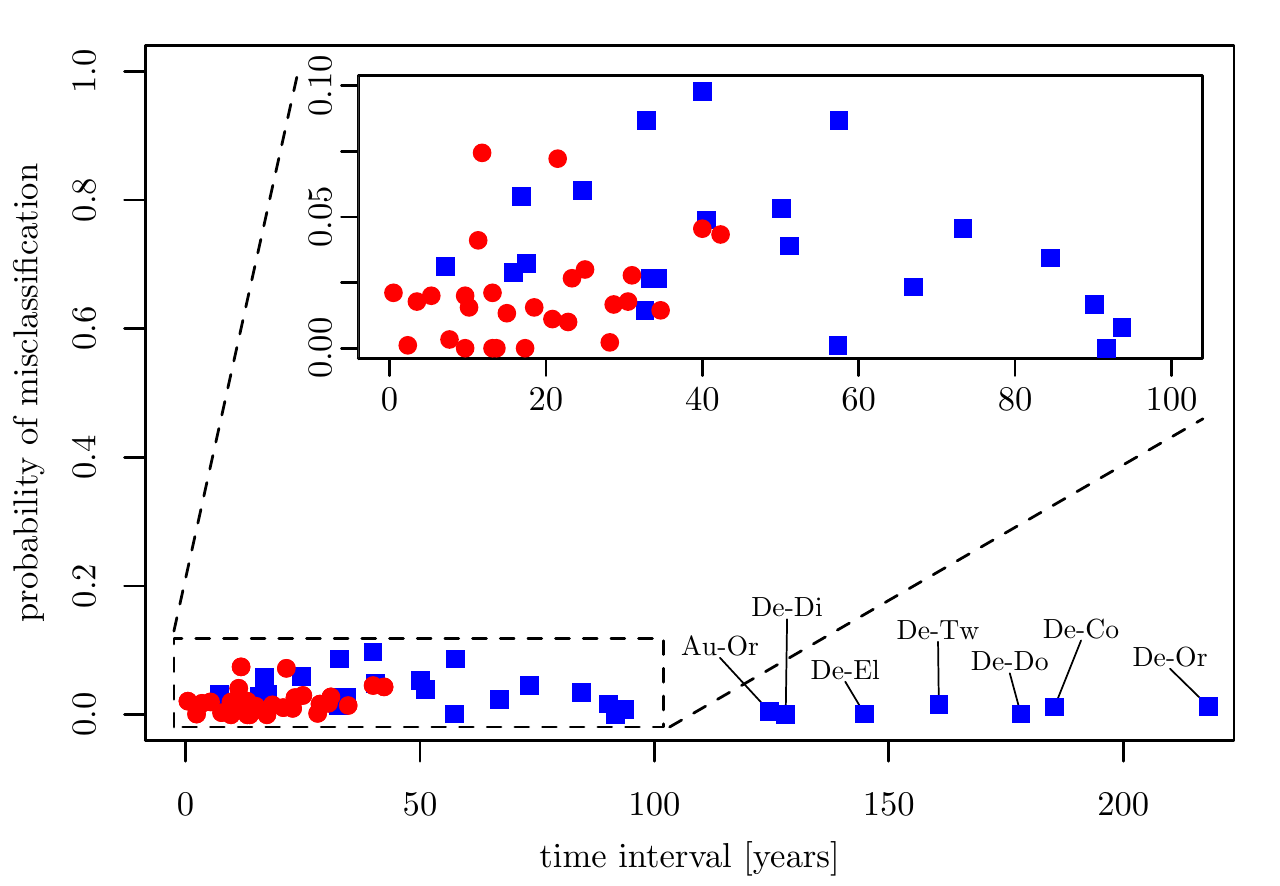}
\caption{The scatterplot of the error rates of pairwise classifications and the time intervals separating classified writers. Each marker denotes one pair of authors - squares for English, and dots for Polish language. All points with time interval greater than 100 years, except one, correspond to pairs in which one author is Daniel Defoe. These points are labeled with writers' names, abbreviated in the same manner as in Tables \ref{tab_clas_global_only_unweighted}, \ref{tab_clas_global_weighted_and_unweighted}, \ref{tab_clas_local}. For the rest of pairs, the scatterplot is presented in more detail in the inset.}
\label{fig_time}
\end{figure}

Another interesting subject is how the classification is affected by the length of the timespan during which the studied books were published. The publication dates of the studied books vary from 1719 to 1949. It is well known that language evolves over time; some aspects of its evolution can be captured by quantitave means and related to the processes taking place in a society and its culture \cite{Michel2010, Lansdall-Welfare2014, Lansdall-Welfare2017}. Such cultural and linguistic changes may possibly affect authorship attribution - it can be suspected that literary works written around the same time are more likely to be confused than these separated by a long time interval. In order to get some preliminary insight into this issue (based on the texts selected here), the following approach is utilized. For each author, a "time centroid" is determined  as an arithmetic mean of publication dates of the works studied in this paper. Then a distance between the centroids (in years) is assigned to each pair of authors and treated as a time interval separating them. Next, for each two authors a classification considering only their works is performed multiple times (feature space consists of weighted degrees and clustering coefficients of the 12 most frequent words; the texts of each author are split into training and test set with ratio 4:2). The error rates in such pairwise classifications (for each pair of authors) are then plotted versus time intervals (Figure \ref{fig_time}) to assess whether the authors separated by greater time intervals are easier to distinguish between or not. At the first glance, it seems that in the case of English, for the writers who are separated by more than 100 years the probability of misclassification is lower than 2\%, while for others it varies more significantly. However, it must be noticed that all except one time interval greater than 100 years pertain to Daniel Defoe, who lived much earlier than the rest of the writers. When considering only the time intervals shorter than 100 years (Figure \ref{fig_time}, inset), no clear relationship between the classification's accuracy and the length of the time interval separating authors can be observed. It must be remembered though, that these results do not pretend to be general - they apply to the set of studied texts, but not necessarily to other sets of literary works; the reliable analysis of the effect of language evolution on authorship attribution, possibly with the proposed approach, would require texts from a larger timespan.

Two additional remarks regarding the obtained results can be made. Firstly, it can be noticed in Figure \ref{fig_accuracy_freq_clustering} that for the Polish books, reaching the limit of the classification accuracy requires less words than for the books in English (it is most visible in the case of classification based on $\widetilde{\text{str}}(v)$ and on both $\widetilde{\text{str}}(v)$ and $\widetilde{C}(v)$). Secondly, the differences between the accuracies of classification based on the weighted network characteristics and the accuracies of classification employing the characteristics in unweighted version, are smaller for the Polish texts than for the English ones. This can be observed in Figure \ref{fig_accuracy_1_char}, for example, for the clustering coefficient.

These effects may perhaps be related to the general structural differences between the languages: while in Polish, like in all Slavic languages, the syntactic role of a word in a sentence is determined mainly by inflection, the English syntax relies more on the word order and on utilizing function words (articles, auxiliary verbs, etc). Function words are among the most frequent ones; one may anticipate that in a language whose grammar imposes stronger conditions on their usage and order, there is less room for the diversity related to an individual writing style. In such a case, more words need to be included into an analysis to obtain reliable accuracy of the authorship classification. Inflection, on the other hand, leads to the increase of the number of words that are treated as distinct - as during the construction of networks the words are not lemmatized. If one constructs the weighted word-adjacency networks from the English and Polish texts samples of equal length, these networks will have the same sum of edges' weights, but the network corresponding to a Polish text will have more vertices. Since word-adjacency networks are connected (every vertex is reachable from any other vertex) and have edge weights expressed by natural numbers, it can be hypothesized that from two weighted networks with the same sum of edges' weights, the network with the larger number of vertices should generally have more low-weight edges and therefore be more similar to an unweighted network. If so, the weighted and the unweighted characteristics of networks should differ less for the Polish texts than for the English ones, as observed. It must be remembered, however, that the presented explanations can only be treated as suppositions; determining the relationship between the two discussed effects and the attributes of the languages, in a reliable way, would require a separate analysis which is beyond the scope of this paper.

\section{Summary}

The presented results confirm the usability of complex networks as a representation of natural language. Though simple in construction and intuitive, they carry within their structure exploitable information on the underlying language sample. Constructing such networks from English and Polish literary texts and studying their properties have led to the possibility of distinguishing individual writing styles. The fact that the presented approach works well for both Polish and English texts suggests that it can probably be treated as applicable to a vast group of languages - as Polish and English are examples of the languages substantially different from each other, in terms of the origin, morphology, and grammar features.

It can be observed that the vectors representing texts belonging to the same author have a noticeable tendency to group together in the space of global network characteristics, both in the case of unweighted and weighted networks. Investigating this tendency, by employing decision tree ensembles to perform the classification of texts with respect to authorship and with network parameters as attributes, results in the classification accuracy higher than the accuracy of a random choice; nevertheless, it is rather questionable to state that the global properties of networks are able to distinguish between authors effectively (at least within the studied set of texts). The networks' property of having different global parameters for texts with different authorship, that is observed for selected groups of authors (Figure \ref{fig_scatterplots}), becomes much less evident as the number of authors grows. This suggests that the global features of word-adjacency networks are to a considerable extent universal for the language and non-specific to individual language users.

However, a division of the set of texts into subsets consisting of works of the same author is possible by analysing the local properties of networks. Calculating the vertex parameters corresponding to the several most frequent words (or punctuation marks), creating a vector of these parameters for each text, and applying the clustering or classification algorithm allows for the evident separation of the texts with respect to their authorship. Among the studied network characteristics, the vertex strength and the weighted clustering coefficient (in an appropriately normalized form) turn out to be the most effective in capturing individual language features, for both the English and the Polish texts. Data clustering in the clustering coefficient's space for only the 12 most frequent words (including punctuation marks) reveals the emergence of text groups that clearly can be related to particular authors. The classification with the decision tree ensembles results in obtaining the average accuracy of around 85-90\%. This confirms that the attributes of selected vertices in a word-adjacency network can be treated to a large extent as typical to an individual style. As these attributes describe a given vertex along with its network neighbourhood, it can be concluded that they are able to capture the specific way the individual authors use particular words and punctuation marks in their writing.

One could argue that because the vertex strength is directly related to the word frequency, and because various local characteristics may be correlated with vertex strength, the method of distinguishing between the authors, based on an analysis of these characteristics, does not differ significantly from the basic approach to authorship attribution, relying on comparing word frequencies. However, it must be remembered that in order to eliminate purely frequency-based effects on the characteristics other than vertex strength, all the studied quantities are normalized. This is done by relating characteristics' values to the average values in the networks corresponding to randomized texts. It may therefore be stated that the analysis of word frequencies and the analysis of network parameters are distinct; also, the fact that network parameters and word frequencies combined lead to the classification results better than the results for any of them studied separately indicates that both these types of text features carry complementary, non-redundant information.

From a practical perspective, the correspondence between the text authorship and the structure of the word-adjacency network seems to be potentially useful when combined with other lexical features of the text. As mentioned above, the accuracy of the classification involving word frequencies (which do not have to be treated as the network parameters) and clustering coefficients (which are purely network-based) achieves the results better than the results obtained from the two methods separately; with such an approach it turns out to be sufficient to select only 4-5 words to get an average of 80\% of the correct classifications. When dealing with the problem of authorship attribution in textual data sets where the number of words available for analysis is somehow limited, it may be profitable to combine the extraction of the basic text features with the network analysis, especially since it requires little preprocessing.

It is worth noting that the presented approach to explore the diversity of language styles is very simple and straightforward - it relies on studying the characteristics of a few most frequent words and punctuation marks. In a future study, it would be recommended to examine the possibility of word selection by the criteria other than frequency, and also to investigate whether there are other features of language that can be captured by studying the structure of linguistic networks.

\nolinenumbers

\bibliographystyle{model5-num-names}

\bibliography{bibliography_file}
\addcontentsline{toc}{section}{References} 

\clearpage
\section*{Appendix\label{appendix}}
\addcontentsline{toc}{section}{Appendix} 

\setcounter{table}{0}
\renewcommand{\thetable}{A.\arabic{table}}

\begin{table}[!ht]
\footnotesize
\centering
\caption{The set of English books studied in this paper. Usual words and punctuation marks are listed separately, although they are treated in the same way during the analysis. Average length of a sentence is the average number of words between punctuation marks denoting the end of a sentence (full stop, question mark, exclamation mark, ellipsis).}

\begin{tabular}{|c|l|c|r|r|r|r|}
\hline
Author & \multicolumn{1}{c|}{Title} & \parbox[c][2.25cm][c]{0.95cm}{\centering Year of publishing} & \parbox[c][][c]{1.15cm}{\centering Number of words (in thousands)} & \parbox[c][][c]{1.15cm}{\centering Number of punctuation marks (in thousands)} & \parbox[c][][c]{1.15cm}{\centering Number of sentences (in thousands)} & \parbox[c][][c]{1.15cm}{\centering Average length of a sentence} \\ \hline \hline \hline
\multirow{6}{*}{\begin{tabular}[c]{@{}l@{}}\parbox[c][2cm][c]{2cm}{\centering   Arthur \\ Conan \\ Doyle \\(1859-1930)}\end{tabular}} & \textit{Micah Clarke} & 1888 & 178.1 & 23.8 & 9.2 & 19.3 \\ \cline{2-7} 
 & \textit{The Adventures of Sherlock Holmes} & 1892 & 104.6 & 15.0 & 6.9 & 15.1 \\ \cline{2-7} 
 & \textit{The Exploits of Brigadier Gerard} & 1896 & 74.7 & 9.8 & 4.0 & 18.6 \\ \cline{2-7} 
 & \textit{The Lost World} & 1912 & 75.8 & 10.2 & 4.5 & 17.0 \\ \cline{2-7} 
 & \textit{The Refugees} & 1893 & 122.9 & 17.3 & 7.8 & 15.8 \\ \cline{2-7} 
 & \textit{The Valley of Fear} & 1915 & 57.7 & 8.3 & 4.3 & 13.5 \\ \hline \hline
\multirow{6}{*}{\begin{tabular}[c]{@{}l@{}}\parbox[c][2cm][c]{2cm}{\centering   Charles \\ Dickens \\ (1812-1870)}\end{tabular}} & \textit{A Tale of Two Cities} & 1859 & 136.2 & 23.2 & 7.8 & 17.5 \\ \cline{2-7} 
 & \textit{Barnaby Rudge} & 1841 & 254.0 & 44.3 & 12.7 & 20.0 \\ \cline{2-7} 
 & \textit{David Copperfield} & 1850 & 356.0 & 63.0 & 19.5 & 18.3 \\ \cline{2-7} 
 & \textit{Oliver Twist} & 1838 & 157.7 & 29.4 & 9.2 & 17.2 \\ \cline{2-7} 
 & \textit{The Mystery of Edwin Drood} & 1870 & 96.0 & 16.6 & 5.7 & 16.9 \\ \cline{2-7} 
 & \textit{The Pickwick Papers} & 1837 & 300.3 & 55.6 & 16.4 & 18.3 \\ \hline \hline
\multirow{6}{*}{\begin{tabular}[c]{@{}l@{}}\parbox[c][2cm][c]{2cm}{\centering   Daniel \\ Defoe \\ (1660-1731)}\end{tabular}} & \textit{Colonel Jack} & 1722 & 141.4 & 20.6 & 4.2 & 33.6 \\ \cline{2-7} 
 & \textit{Memoirs of a Cavalier} & 1720 & 101.2 & 13.8 & 2.6 & 39.5 \\ \cline{2-7} 
 & \textit{Roxana: The Fortunate Mistress} & 1724 & 160.9 & 23.4 & 3.8 & 41.9 \\ \cline{2-7} 
 & \textit{Moll Flanders} & 1722 & 136.2 & 18.9 & 3.2 & 42.1 \\ \cline{2-7} 
 & \textit{Robinson Crusoe} & 1719 & 232.3 & 34.0 & 4.1 & 56.6 \\ \cline{2-7} 
 & \textit{Captain Singleton} & 1720 & 110.8 & 16.1 & 2.4 & 47.0 \\ \hline \hline
\multirow{6}{*}{\begin{tabular}[c]{@{}l@{}}\parbox[c][2cm][c]{2cm}{\centering   George \\ Eliot \\(1819-1880)}\end{tabular}} & \textit{Adam Bede} & 1859 & 215.1 & 28.0 & 9.0 & 24.0 \\ \cline{2-7} 
 & \textit{Daniel Deronda} & 1876 & 311.1 & 39.2 & 14.3 & 21.7 \\ \cline{2-7} 
 & \textit{Felix Holt, the Radical} & 1866 & 182.2 & 24.1 & 8.1 & 22.6 \\ \cline{2-7} 
 & \textit{Middlemarch} & 1872 & 318.1 & 41.2 & 14.9 & 21.3 \\ \cline{2-7} 
 & \textit{Romola} & 1863 & 227.9 & 29.5 & 9.1 & 25.0 \\ \cline{2-7} 
 & \textit{The Mill on the Floss} & 1860 & 207.3 & 30.5 & 9.0 & 23.1 \\ \hline \hline
\multirow{6}{*}{\begin{tabular}[c]{@{}l@{}}\parbox[c][2cm][c]{2cm}{\centering   George \\ Orwell \\ (1903-1950)}\end{tabular}} & \textit{Animal Farm} & 1945 & 30.1 & 3.8 & 1.6 & 18.3 \\ \cline{2-7} 
 & \textit{Burmese Days} & 1934 & 97.7 & 15.1 & 7.4 & 13.2 \\ \cline{2-7} 
 & \textit{Coming up for Air} & 1939 & 82.8 & 10.4 & 5.3 & 15.7 \\ \cline{2-7} 
 & \textit{Down and Out in Paris and London} & 1933 & 66.8 & 9.9 & 4.0 & 16.5 \\ \cline{2-7} 
 & \textit{Keep the Aspidistra Flying} & 1936 & 87.1 & 14.3 & 7.8 & 11.2 \\ \cline{2-7} 
 & \textit{Nineteen Eighty-Four} & 1949 & 103.7 & 14.2 & 6.7 & 15.5 \\ \hline \hline
\multirow{6}{*}{\begin{tabular}[c]{@{}l@{}}\parbox[c][2cm][c]{2cm}{\centering   Jane \\ Austen \\ (1775-1817)}\end{tabular}} & \textit{Emma} & 1815 & 160.3 & 26.5 & 8.6 & 18.5 \\ \cline{2-7} 
 & \textit{Mansfield Park} & 1814 & 159.8 & 22.5 & 6.9 & 23.0 \\ \cline{2-7} 
 & \textit{Northanger Abbey} & 1818 & 77.3 & 11.4 & 3.6 & 21.4 \\ \cline{2-7} 
 & \textit{Persuasion} & 1818 & 83.3 & 12.4 & 3.7 & 22.8 \\ \cline{2-7} 
 & \textit{Pride and Prejudice} & 1813 & 121.8 & 17.2 & 6.0 & 20.3 \\ \cline{2-7} 
 & \textit{Sense and Sensibility} & 1811 & 119.5 & 18.0 & 5.2 & 23.0 \\ \hline \hline
\multirow{6}{*}{\begin{tabular}[c]{@{}l@{}}\parbox[c][2cm][c]{2cm}{\centering   Joseph \\ Conrad \\ (1857-1924)}\end{tabular}} & \textit{An Outcast of the Islands} & 1896 & 104.4 & 18.0 & 8.8 & 11.9 \\ \cline{2-7} 
 & \textit{Chance: A Tale in Two Parts} & 1913 & 137.4 & 18.3 & 9.6 & 14.4 \\ \cline{2-7} 
 & \textit{Lord Jim} & 1900 & 129.3 & 20.4 & 8.9 & 14.5 \\ \cline{2-7} 
 & \textit{Nostromo: A Tale of the Seaboard} & 1904 & 168.3 & 24.3 & 10.1 & 16.7 \\ \cline{2-7} 
 & \textit{Under Western Eyes} & 1911 & 112.8 & 16.9 & 8.6 & 13.1 \\ \cline{2-7} 
 & \textit{Victory: An Island Tale} & 1915 & 114.8 & 18.3 & 8.6 & 13.4 \\ \hline \hline
\multirow{6}{*}{\begin{tabular}[c]{@{}l@{}}\parbox[c][2cm][c]{2cm}{\centering   Mark \\ Twain \\ (1835-1910)}\end{tabular}} & \textit{Following the Equator} & 1897 & 186.7 & 26.1 & 9.2 & 20.3 \\ \cline{2-7} 
 & \textit{Life on the Mississippi} & 1883 & 143.9 & 21.1 & 6.8 & 21.2 \\ \cline{2-7} 
 & \textit{The Adventures of Huckleberry Finn} & 1884 & 110.8 & 16.9 & 5.9 & 18.8 \\ \cline{2-7} 
 & \textit{The Adventures of Tom Sawyer} & 1876 & 70.5 & 11.6 & 4.9 & 14.4 \\ \cline{2-7} 
 & \textit{The Innocents Abroad} & 1869 & 192.4 & 25.4 & 8.8 & 22.0 \\ \cline{2-7} 
 & \textit{The Prince and the Pauper} & 1882 & 67.2 & 10.8 & 3.3 & 20.1 \\ \hline
\end{tabular}
\label{tab_list_of_texts_en}
\end{table}

\begin{table}[!ht]
\footnotesize
\centering
\caption{The set of Polish books studied in this paper. Usual words and punctuation marks are listed separately, although they are treated in the same way during the analysis. Average length of a sentence is the average number of words between punctuation marks denoting the end of a sentence (full stop, question mark, exclamation mark, ellipsis).}

\begin{tabular}{|c|l|c|r|r|r|r|}
\hline
Author & \multicolumn{1}{c|}{Title} & \parbox[c][2.25cm][c]{0.95cm}{\centering Year of publishing} & \parbox[c][][c]{1.15cm}{\centering Number of words (in thousands)} & \parbox[c][][c]{1.15cm}{\centering Number of punctuation marks (in thousands)} & \parbox[c][][c]{1.15cm}{\centering Number of sentences (in thousands)} & \parbox[c][][c]{1.15cm}{\centering Average length of a sentence} \\ \hline \hline \hline
\multirow{6}{*}{\parbox[c][2cm][c]{2cm}{\centering Bolesław \\ Prus \\ (1847-1912)}} & \textit{Anielka} & 1885 & 46.6 & 11.6 & 5.7 & 8.2 \\ \cline{2-7} 
 & \textit{Dzieci} & 1909 & 66.1 & 19.1 & 9.1 & 7.3 \\ \cline{2-7} 
 & \textit{Emancypantki} & 1894 & 249.2 & 63.5 & 30.2 & 8.3 \\ \cline{2-7} 
 & \textit{Faraon} & 1897 & 193.3 & 45.8 & 19.7 & 9.8 \\ \cline{2-7} 
 & \textit{Lalka} & 1890 & 246.4 & 61.2 & 28.2 & 8.7 \\ \cline{2-7} 
 & \textit{Placówka} & 1886 & 68.6 & 17.5 & 7.7 & 8.9 \\ \hline \hline
\multirow{6}{*}{\parbox[c][2cm][c]{2cm}{\centering Eliza \\ Orzeszkowa \\ (1841-1910)}} & \textit{Cham} & 1888 & 59.4 & 14.2 & 4.4 & 13.4 \\ \cline{2-7} 
 & \textit{Dziurdziowie} & 1885 & 50.6 & 10.8 & 3.7 & 13.8 \\ \cline{2-7} 
 & \textit{Jędza} & 1891 & 40.1 & 9.9 & 3.7 & 11.0 \\ \cline{2-7} 
 & \textit{Marta} & 1873 & 62.1 & 12.5 & 4.5 & 13.8 \\ \cline{2-7} 
 & \textit{Meir Ezofowicz} & 1878 & 97.3 & 18.4 & 6.9 & 14.1 \\ \cline{2-7} 
 & \textit{Nad Niemnem} & 1888 & 164.4 & 32.6 & 11.8 & 14.0 \\ \hline \hline
\multirow{6}{*}{\parbox[c][2cm][c]{2cm}{\centering Henryk \\ Sienkiewicz \\ (1846-1916)}} & \textit{Ogniem i mieczem} & 1884 & 239.0 & 51.9 & 19.4 & 12.3 \\ \cline{2-7} 
 & \textit{Potop} & 1886 & 380.4 & 84.7 & 32.8 & 11.6 \\ \cline{2-7} 
 & \textit{Quo Vadis} & 1896 & 168.3 & 34.1 & 11.4 & 14.7 \\ \cline{2-7} 
 & \textit{Rodzina Połanieckich} & 1894 & 210.2 & 47.3 & 15.1 & 13.9 \\ \cline{2-7} 
 & \textit{W pustyni i w puszczy} & 1912 & 99.5 & 18.1 & 6.5 & 15.2 \\ \cline{2-7} 
 & \textit{Wiry} & 1910 & 92.7 & 20.0 & 6.6 & 14.0 \\ \hline \hline
\multirow{6}{*}{\parbox[c][2cm][c]{2cm}{\centering Jan \\ Lam \\ (1838-1886)}} & \textit{Dziwne karyery} & 1881 & 70.9 & 14.8 & 4.7 & 15.2 \\ \cline{2-7} 
 & \textit{Humoreski} & 1883 & 37.6 & 7.5 & 2.3 & 16.4 \\ \cline{2-7} 
 & \textit{Idealiści} & 1876 & 100.9 & 21.3 & 7.0 & 14.5 \\ \cline{2-7} 
 & \textit{Koroniarz w Galicyi} & 1870 & 62.0 & 11.6 & 3.3 & 18.9 \\ \cline{2-7} 
 & \textit{Rozmaitości i powiastki} & 1878 & 52.4 & 10.3 & 2.6 & 20.0 \\ \cline{2-7} 
 & \textit{Wielki świat Capowic} & 1869 & 37.7 & 6.7 & 1.6 & 23.0 \\ \hline \hline
\multirow{6}{*}{\parbox[c][2cm][c]{2cm}{\centering Janusz \\ Korczak \\ (1879-1942)}} & \textit{Bankructwo małego Dżeka} & 1924 & 41.7 & 10.3 & 4.3 & 9.6 \\ \cline{2-7} 
 & \textit{Dzieci ulicy} & 1901 & 50.2 & 13.2 & 5.9 & 8.5 \\ \cline{2-7} 
 & \textit{Dziecko salonu} & 1906 & 54.1 & 16.7 & 6.3 & 8.6 \\ \cline{2-7} 
 & \textit{Kajtuś czarodziej} & 1934 & 46.0 & 17.9 & 9.8 & 4.7 \\ \cline{2-7} 
 & \textit{Król Maciuś na wyspie bezludnej} & 1923 & 45.9 & 12.1 & 5.2 & 8.8 \\ \cline{2-7} 
 & \textit{Król Maciuś Pierwszy} & 1923 & 67.0 & 15.1 & 6.2 & 10.8 \\ \hline \hline
\multirow{6}{*}{\parbox[c][2cm][c]{2cm}{\centering Józef \\ Ignacy \\ Kraszewski \\ (1812-1887)}} & \textit{Barani Kożuszek} & 1898 & 60.3 & 14.8 & 5.5 & 11.0 \\ \cline{2-7} 
 & \textit{Boża opieka} & 1873 & 51.3 & 11.9 & 4.9 & 10.5 \\ \cline{2-7} 
 & \textit{Boży gniew} & 1886 & 112.0 & 24.9 & 7.8 & 14.4 \\ \cline{2-7} 
 & \textit{Bracia rywale} & 1877 & 42.0 & 10.5 & 3.8 & 11.1 \\ \cline{2-7} 
 & \textit{Infantka} & 1884 & 111.3 & 23.2 & 8.2 & 13.6 \\ \cline{2-7} 
 & \textit{Złote jabłko} & 1853 & 107.5 & 22.6 & 9.7 & 11.1 \\ \hline \hline
\multirow{6}{*}{\parbox[c][2cm][c]{2cm}{\centering Stefan \\ Żeromski \\ (1846-1925)}} & \textit{Dzieje grzechu} & 1908 & 148.5 & 35.0 & 15.9 & 9.3 \\ \cline{2-7} 
 & \textit{Ludzie bezdomni} & 1900 & 97.2 & 20.2 & 8.6 & 11.3 \\ \cline{2-7} 
 & \textit{Popioły} & 1902 & 222.4 & 46.5 & 21.0 & 10.6 \\ \cline{2-7} 
 & \textit{Przedwiośnie} & 1924 & 81.3 & 17.5 & 7.3 & 11.2 \\ \cline{2-7} 
 & \textit{Syzyfowe prace} & 1897 & 61.9 & 11.6 & 4.5 & 13.7 \\ \cline{2-7} 
 & \textit{Wierna rzeka} & 1912 & 39.5 & 8.6 & 4.0 & 9.8 \\ \hline \hline
\multirow{6}{*}{\parbox[c][2cm][c]{2cm}{\centering Władysław \\ Reymont \\ (1867-1925)}} & \textit{Bunt} & 1924 & 40.2 & 9.0 & 3.8 & 10.6 \\ \cline{2-7} 
 & \textit{Fermenty} & 1897 & 117.9 & 30.9 & 9.8 & 12.0 \\ \cline{2-7} 
 & \textit{Lili} & 1899 & 25.0 & 6.3 & 2.2 & 11.6 \\ \cline{2-7} 
 & \textit{Marzyciel} & 1910 & 47.7 & 11.9 & 4.4 & 10.8 \\ \cline{2-7} 
 & \textit{Wampir} & 1911 & 53.0 & 13.0 & 4.5 & 11.7 \\ \cline{2-7} 
 & \textit{Ziemia obiecana} & 1899 & 170.9 & 38.0 & 13.3 & 12.9 \\ \hline
\end{tabular}

\label{tab_list_of_texts_pl}
\end{table}

\end{document}